\pdfoutput=1
\documentclass[twoside,twocolumn,10pt]{article}

\usepackage{wscg}           
\RequirePackage{ifpdf}
\ifpdf
 \RequirePackage[pdftex]{graphicx}
 \RequirePackage[pdftex]{color}
\else
 \RequirePackage[dvips,draft]{graphicx}
 \RequirePackage[dvips]{color}
\fi

\usepackage{cite}

\usepackage{amssymb}
\usepackage{bm} 
\usepackage{amsmath}

\usepackage{booktabs}
\usepackage{multirow}
\usepackage{array}  

\usepackage{pdfpages}
\usepackage{caption}
\usepackage{subcaption}

\usepackage[table]{xcolor}
\usepackage{tabularx}

\newcommand{\cellimg}[1]{\includegraphics[width=\linewidth]{#1}}
\newcolumntype{C}{>{\centering\arraybackslash}X}
\newcolumntype{L}{>{\raggedright\arraybackslash}X}

\usepackage{nopageno}

\usepackage{url}
\makeatletter
\def\Uslash{\mathbin{\mathchar`\/}\@ifnextchar{/}{\kern-.15em}{}}
\g@addto@macro\UrlSpecials{\do \/ {\Uslash}}
\def\Ucolon{\mathbin{\mathchar`:}\@ifnextchar{/}{\kern-.1em}{}}
\g@addto@macro\UrlSpecials{\do : {\Ucolon}}
\makeatother

\title{VisibilitySplat: Ray-Traced Occlusion for Relighting with 3D Gaussian Splatting}

\author{
\parbox{0.4\textwidth}{\centering
Sven Kluge\\[1mm]
University of Rostock\\
Albert-Einstein-Stra{\ss}e 22\\
18059 Rostock, Germany \\[1mm]
sven.kluge@uni-rostock.de
}
\hspace{0.05\textwidth}
\parbox{0.4\textwidth}{\centering
Oliver Staadt\\[1mm]
University of Rostock\\
Albert-Einstein-Stra{\ss}e 22\\
18059 Rostock, Germany \\[1mm]
oliver.staadt@uni-rostock.de
}
}

\begin{document}

\twocolumn[{\csname @twocolumnfalse\endcsname

\maketitle

\begin{abstract}
\noindent
We address inverse rendering with 3D Gaussian Splatting (3DGS) by introducing explicit visibility modeling for relighting as well as material and light decomposition. Our approach augments the standard rasterized 3DGS pipeline with a sparse visibility oracle that estimates transmittance-weighted occlusion for Gaussian primitives from sampled lighting directions. This oracle is implemented using a point-based ray tracer accelerated by a bounding volume hierarchy (BVH) and queried sparsely within the inverse-rendering objectives, while the original 3DGS loop is retained for training efficiency, interactive visualization, and tooling compatibility. Shading follows a physically based rendering (PBR) formulation with a principled microfacet bidirectional reflectance distribution function (BRDF) and low-resolution high-dynamic-range (HDR) environment maps. We evaluate our method on the TensoIR synthetic dataset against baselines based on neural radiance fields (NeRFs), tensor representations, and 3DGS. The results show competitive performance and improvements over several earlier baselines for novel-view synthesis (NVS), albedo reconstruction, and relighting. Ablations indicate that explicit visibility modeling is an important contributor to relighting and material-illumination disentanglement, while pure PBR supervision can favor appearance fitting in NVS. Overall, the findings suggest that sparse, point-based visibility queries effectively complement rasterized splatting for inverse rendering while preserving the practical advantages of the 3DGS ecosystem.
\end{abstract}

\subsection*{Keywords}
inverse rendering; 3D Gaussian splatting; relighting; visibility; physically based rendering

\vspace*{1.0\baselineskip}
}]

\section{Introduction}

\copyrightspace

\begin{figure*}
\centering
\includegraphics{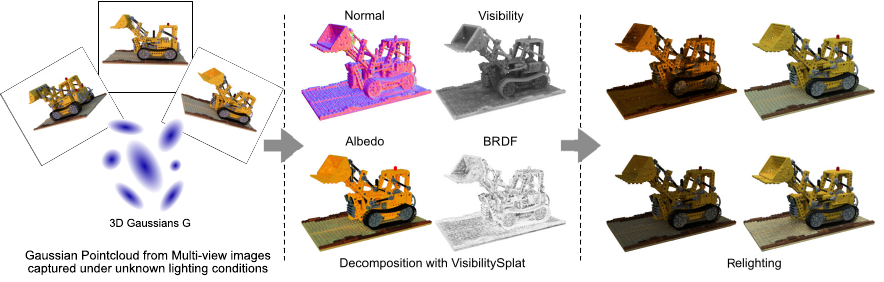}
\caption{\textbf{VisibilitySplat.} Given a Gaussian point cloud optimized from multi-view images under unknown illumination (left), VisibilitySplat recovers physical properties (normals, materials, and illumination) using a visibility ray tracer (center), enabling relighting (right).}
\label{fig:teaser}
\end{figure*}

Inverse rendering (IR) seeks to recover geometry, reflectance, and illumination from images by inverting the image formation process. Neural radiance fields (NeRFs) \cite{mildenhall:2021:nerf} enable high-fidelity novel-view synthesis (NVS) and have been extended toward relightable decompositions under unknown illumination (e.g., NeRFactor \cite{zhang:2021:nerfactor} and TensoIR \cite{jin:2023:tensoir}), but their implicit volumetric formulation is costly to train and render. 3D Gaussian Splatting (3DGS) \cite{kerbl:2023:3dgs} instead optimizes explicit point primitives via differentiable rasterization, enabling real-time rendering and practical tooling, and has recently been adopted for IR (e.g., 3D Gaussian Splatting for Inverse Rendering (GS-IR) \cite{liang:2024:gs-ir} and Gaussian Inverse Rendering (GIR) \cite{shi:2025:gir}).
A key challenge is visibility: screen-space alpha compositing provides limited information about cast shadows and mutual occlusion, which are crucial for stabilizing material-light disentanglement under unknown illumination. While 3DGS-based IR pipelines benefit from fast forward rendering, visibility signals derived from the rasterizer alone can be too weak or too view-dependent to constrain shadows robustly. Recent work, including Inter-Reflective Gaussian Splatting (IRGS), therefore augments splatting with ray-based visibility estimation \cite{gao:2024:relightable3dgaussians, gu:2025:irgs}.
We take a pragmatic step along this trajectory by adding a sparse visibility oracle that estimates transmittance-weighted occlusion from sampled lighting directions, while retaining the original 3DGS rasterization loop for training efficiency and interactive visualization. This lightweight division of labor injects physically meaningful visibility constraints without replacing the splatting renderer (Fig.~\ref{fig:teaser}).
Rather than proposing a highly specialized approximation or learned visibility model, we deliberately study a direct formulation based on explicit visibility queries. This provides a reference baseline for assessing how much transmittance-weighted visibility contributes to IR and relighting with 3DGS. Our contributions are as follows:
(i) a sparse visibility oracle for 3DGS that estimates how much light reaches each Gaussian from sampled environment directions, implemented with a point-based ray tracer accelerated by a bounding volume hierarchy (BVH);
(ii) a reference baseline on the TensoIR synthetic benchmark, including comparisons to NeRF-based, tensor-based, and 3DGS-based IR methods and ablations that isolate the effect of explicit visibility.

\begin{figure*}
\centering
\includegraphics[width=\textwidth]{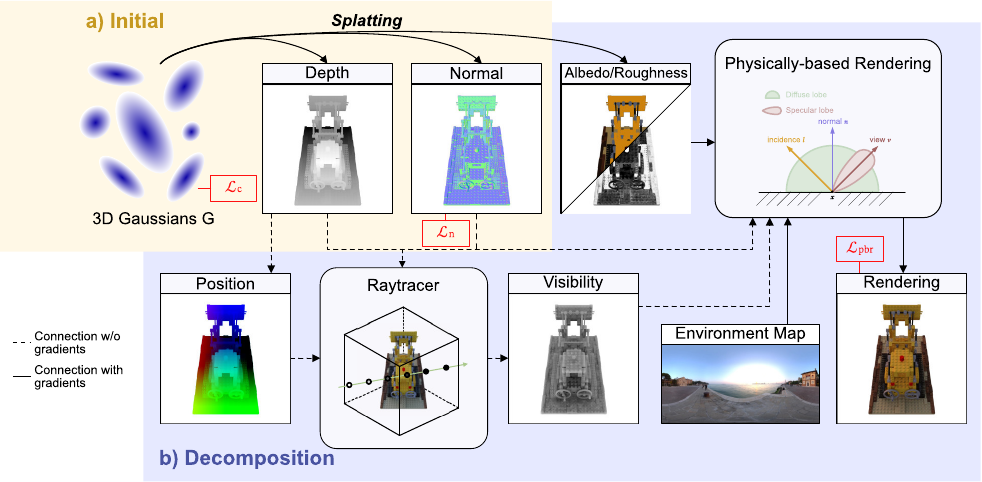}
\caption{Method Overview. VisibilitySplat is a two-stage, 3DGS-based IR pipeline that reconstructs scene geometry, materials, and environment illumination from multi-view images. Stage 1 pretrains a 3D Gaussian point cloud with the standard 3DGS attributes and per-splat normals derived from depth. Stage 2 adds a sparse visibility oracle, implemented with BVH-accelerated point-based ray tracing, to compute transmittance-weighted visibility over the Gaussians. Together with physically based rendering (PBR) using a Disney ``principled'' BRDF \cite{burley:2012:physically}, this enables joint decomposition into material parameters and estimation of a compact high-dynamic-range (HDR) environment map for relighting.}
\label{fig:overview}
\end{figure*}

\section{Related Work}
\subsection{Radiance Fields}
The original NeRF method \cite{mildenhall:2021:nerf} introduced neural volumetric representations for NVS, with subsequent work improving anti-aliasing and efficiency via scale-aware integration and more explicit encodings \cite{barron:2021:mip, barron:2022:mip360, barron:2023:zip, muller:2022:instantngp, fridovich:2022:plenoxels,fridovich:2023:kplanes, chen:2022:tensorf}. 3DGS \cite{kerbl:2023:3dgs} further shifts computation to explicit anisotropic Gaussian primitives rendered by a custom rasterizer, enabling real-time performance. Geometry-oriented variants \cite{huang:2024:2dgs, yu:2024:gaussian} and BVH-based Gaussian ray tracing \cite{loccoz:2024:3dgrt} broaden the representation toward editing and secondary-ray effects.
\subsection{Inverse Rendering}
\textbf{Neural fields for inverse rendering:} IR aims to recover geometry, materials, and illumination from images and is intrinsically ill-posed. Early NeRF-based methods reduced ambiguity with simplifying assumptions: PhySG \cite{zhang:2021:physg} represents bidirectional reflectance distribution functions (BRDFs) and lighting using spherical Gaussians (SGs) but omits complex visibility and indirect transport. NeRFactor \cite{zhang:2021:nerfactor} factorizes shape and reflectance under unknown lighting. InvRender \cite{zhang:2022:invrender} derives spatially varying indirect illumination directly from the learned radiance field to improve albedo recovery. Ref-NeRF \cite{verbin:2022:ref} improves specular handling via a reflected-radiance parameterization that also yields better normals. To raise physical fidelity without losing efficiency, TensoIR \cite{jin:2023:tensoir} builds on tensor-factorized fields to support on-the-fly visibility queries and second-bounce effects during reconstruction. Complementary surface-aligned fields stabilize geometry and normals for IR pipelines \cite{yariv:2021:volsdf, wang:2021:neus}. More recent work targets illumination modeling itself: ENVIDR \cite{liang:2023:envidr} learns a neural environment for shiny scenes. NeRF-as-Emitter \cite{ling:2024:nerf-as-emitter} replaces the distant-light assumption with a near-field light NeRF, improving accuracy under spatially varying illumination, and NeILF++ \cite{zhang:2023:neilf++} jointly learns incident-light and outgoing-radiance fields with inter-reflections to better disentangle material and lighting.

\textbf{Gaussian splatting for inverse rendering:} Adopting 3DGS for IR leverages real-time forward rendering but introduces new challenges: splats lack native surface normals, and forward splatting does not natively model occlusion or multi-bounce light. GS-IR \cite{liang:2024:gs-ir} addresses these by regularizing per-Gaussian normals and baking visibility for secondary effects while recovering geometry, spatially varying bidirectional reflectance distribution functions (SVBRDFs), and environment lighting under unknown illumination. Subsequent work deepens physical correctness: GaussianShader \cite{jiang:2024:gaussianshader} estimates robust normals and per-splat shading for reflective scenes. DeferredGS \cite{wu:2024:deferredgs} decouples geometry from lighting and performs deferred shading for editable relighting and Ref-Gaussian \cite{yao:2024:reflective} targets glossy reflections with a specialized deferred pipeline. Beyond perfect mirrors, SSS-GS \cite{dihlmann:2024:subsurface} introduces subsurface scattering to a 3DGS framework using an explicit surface plus an implicit scattering component. To improve general relighting, RNG \cite{fan:2025:rng} conditions radiance on view and light directions with shadow cues and depth refinement, while BiGS \cite{liu:2024:bigs} unifies surface and volumetric materials via bidirectional spherical harmonics (SH), supporting near- and distant-light setups. Finally, GI-GS \cite{chen:2024:gi-gs} explicitly decomposes direct/indirect components and uses lightweight path tracing on a G-buffer for multi-bounce global illumination, bringing 3DGS-based IR closer to physically faithful relighting at interactive rates. These efforts are complementary to geometry-focused 2D Gaussian Splatting (2DGS) \cite{huang:2024:2dgs} and surface extraction with Gaussian Opacity Fields (GOF) \cite{yu:2024:gaussian}, as well as to BVH-accelerated Gaussian ray tracing for secondary-ray phenomena \cite{loccoz:2024:3dgrt}.

\section{Overview}

\subsection{Gaussian Splatting}
3DGS~\cite{kerbl:2023:3dgs} is a method for rendering 3D scenes in real time, offering high visual quality and fast training. A scene is represented as a cloud of anisotropic 3D Gaussians:
\begin{equation}
g_i(x)=e^{-\frac{1}{2}(x-\mu_i)^T\Sigma^{-1}_i(x-\mu_i)}
\end{equation}
with mean $\mu_i \in \mathbb{R}^3$ and covariance $\Sigma_i$.

The complete parameter set for each Gaussian is defined as $G_i = (\mu_i,\Sigma_i,\alpha_i,c_i)$, where the mean $\mu_i \in \mathbb{R}^3$ specifies the position, the full covariance $\Sigma_i \in \mathbb{R}^{3\times3}$ encodes the ellipsoidal footprint, $\alpha_i$ denotes the opacity, and $c_i$ is the color, stored as a set of SH coefficients. The covariance matrix $\Sigma_i$ can be decomposed into a scaling matrix $\mathbf{S}_i = \mathrm{diag}(s_x, s_y, s_z)$ and a rotation matrix $\mathbf{R}_i$ constructed from a unit quaternion $\mathbf{q}$, which results in $\Sigma_i = \mathbf{R}\mathbf{S}\mathbf{S}^T\mathbf{R}^T$.

With the extrinsic camera matrix $\mathbf{T}$ and intrinsic camera matrix $\mathbf{K}$, the screen-space mean $\mu'_i$ and covariance matrix $\Sigma'$ are defined as
\begin{equation}
     \mu'_i = \mathbf{KT}[\mu_i,1]^T, \quad \Sigma' = \mathbf{JT}\Sigma\mathbf{T}^T\mathbf{J}^T
\end{equation}
where $\mathbf{J}$ is the Jacobian matrix of the affine approximation of the perspective projection~\cite{zwicker:2002:ewa}.

The final color at pixel $u$ is obtained by depth-sorting Gaussians and accumulating color in a front-to-back manner:
\begin{equation}
    \hat{C} = \sum_{i\in \mathbb{N}}T_i g_i\alpha_ic_i, \quad T_i=\prod_{j=1}^{i-1}(1-g_j\alpha_j)
\end{equation}
where $T_i$ denotes the accumulated transmittance, and for readability we use the shorthand $g_i \equiv g_i(u\mid\mu'_i,\Sigma'_i)$.

\subsection{Physically Based Rendering}
We follow a physically based rendering (PBR) formulation similar to recent IR systems such as TensoIR \cite{jin:2023:tensoir}, NeRFactor \cite{zhang:2021:nerfactor}, and GS-IR \cite{liang:2024:gs-ir}. Image formation is expressed by the surface rendering equation with geometric visibility and an energy-conserving microfacet BRDF, while geometry, reflectance, and illumination are represented by differentiable fields and optimized from multi-view images. Formally,
\begin{equation}
    L_o(\mathbf{x},\omega_o)=\int_\Omega f_r(\mathbf{x}, \omega_i, \omega_o)L_i(\mathbf{x},\omega_i)V(\mathbf{x},\omega_i)(\mathbf{n}\cdot\omega_i)d\omega_i.
\end{equation}
Here, $\mathbf{x}$ is the shaded surface point, $\omega_o$ is the outgoing view direction, and $\omega_i$ is an incident lighting direction integrated over the hemisphere $\Omega$ around the surface normal $\mathbf{n}$. $L_o$ and $L_i$ denote outgoing and incident radiance, $f_r$ is the BRDF, $V\in[0,1]$ is the visibility or transmittance term, and $(\mathbf{n}\cdot\omega_i)$ is the cosine foreshortening factor.

For reflectance, we adopt a reduced principled microfacet model in the Disney family \cite{burley:2012:physically} and remove the metallic degree of freedom to keep the inverse problem better constrained. We use a fixed dielectric base reflectance $F_0=0.04$, and write
\begin{equation}
\begin{aligned}
    f_r &=f_{diff}+f_{spec},\\
    f_{spec} &=\frac{D(\mathbf{h})G(\omega_i,\omega_o)F(\omega_o\cdot\mathbf{h};F_0=0.04)}{4(\mathbf{n}\cdot\omega_i)(\mathbf{n}\cdot\omega_o)}
\end{aligned}
\end{equation}
where $D$ is the normal-distribution function, $G$ is the masking-shadowing term, and $F$ is the Fresnel term (typically Schlick) evaluated with fixed $F_0$. Roughness controls the lobe width. Fixing $F_0$ and removing metallicity reduces the number of learnable appearance parameters, which improves stability in this under-constrained setting while retaining a physically plausible BRDF.

In practice, illumination is computed by evaluating the integral over scene lights and/or an environment map together with the visibility term $V$. Many IR systems employ this same BRDF formulation but differ in how geometry, visibility, and indirect effects are estimated. For example, TensoIR introduces efficient online visibility and secondary shading, NeRFactor learns a visibility field, and GS-IR attaches BRDF parameters to Gaussians with an occlusion approximation.

\subsection{Environmental Lighting}
Accurate recovery of incident illumination is essential for robust IR. We model distant lighting as a high-dynamic-range (HDR) spherical environment map parameterized in equirectangular coordinates~\cite{debevec:1998:rendering}. Unlike SGs or mixtures of SGs, this image-based representation preserves high-frequency angular detail and thus supports sharp cast shadows without requiring many analytic lobes.

Each HDR environment-map texel stores radiance $L$, and we optimize these values directly during training with solid-angle compensation to account for the non-uniform texel area of equirectangular maps.

Because reflectance acts as an angular low-pass filter, the recoverable illumination bandwidth is limited by the object's BRDF lobes~\cite{ramamoorthi:2001:AnER}. Accordingly, we use a compact environment-map parameterization, with concrete settings summarized in the implementation details paragraph.

\subsection{Depth and Normal Estimation}
Reconstructing scene geometry, in particular surface normals, is essential for IR and PBR.

\textbf{Depth estimation:} We estimate depth with the expected depth $z_E$ of the Gaussian mixture along each ray, obtained by linearly accumulating depth samples $z_i$ and normalizing by the accumulated weight:
\begin{equation}
    z_E = \frac{\sum_{i \in \mathbb{N}} w_i z_i}{W}, \; W=\sum_{i \in \mathbb{N}} w_i, \; w_i=T_{i}g_i\alpha_i.
\end{equation}
We evaluate alternative depth formulations, including inverse-depth variants, and find that expected depth provides the best overall reconstruction quality and the most stable downstream supervision in our setting. We therefore use $z_E$ as the geometric signal for pseudo-normal supervision in optimization and for computing surface positions that initialize ray-tracing queries.

\begin{figure*}[t]
\centering
\setlength{\tabcolsep}{3pt}            
\renewcommand{\arraystretch}{0.85}     

\resizebox{\textwidth}{!}{%
\scriptsize
\begin{tabular}{@{}l *{8}{>{\centering\arraybackslash}m{0.09\linewidth}} @{}}
    & \textbf{Normal} & \textbf{Albedo} & \textbf{Env.\ Map } 
    & \textbf{Normal} & \textbf{Albedo} & \textbf{Env.\ Map } \\[2pt]
    \midrule

    \textbf{TensoIR}
    & \cellimg{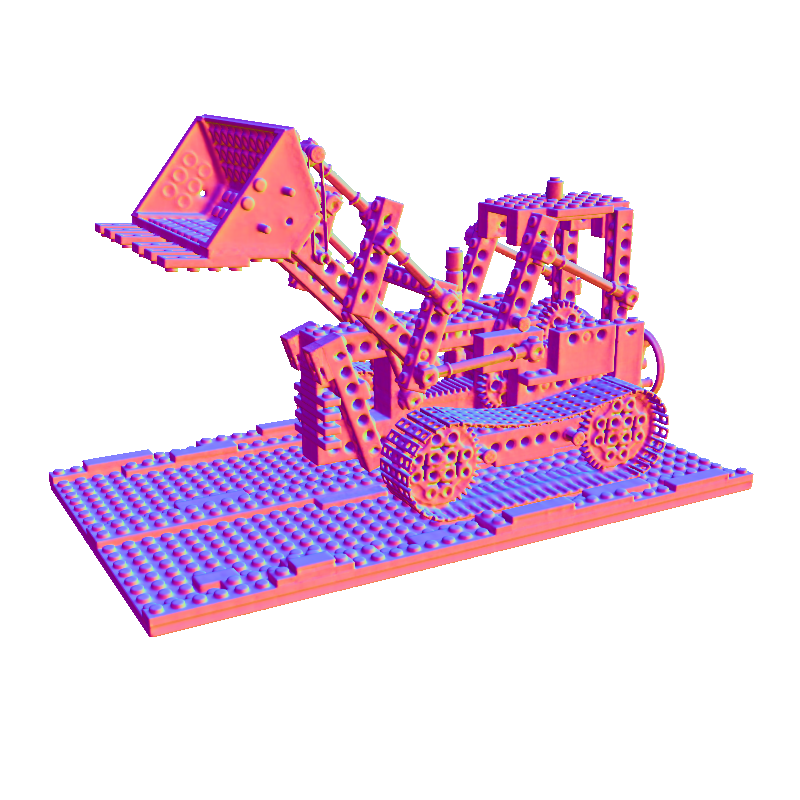}
    & \cellimg{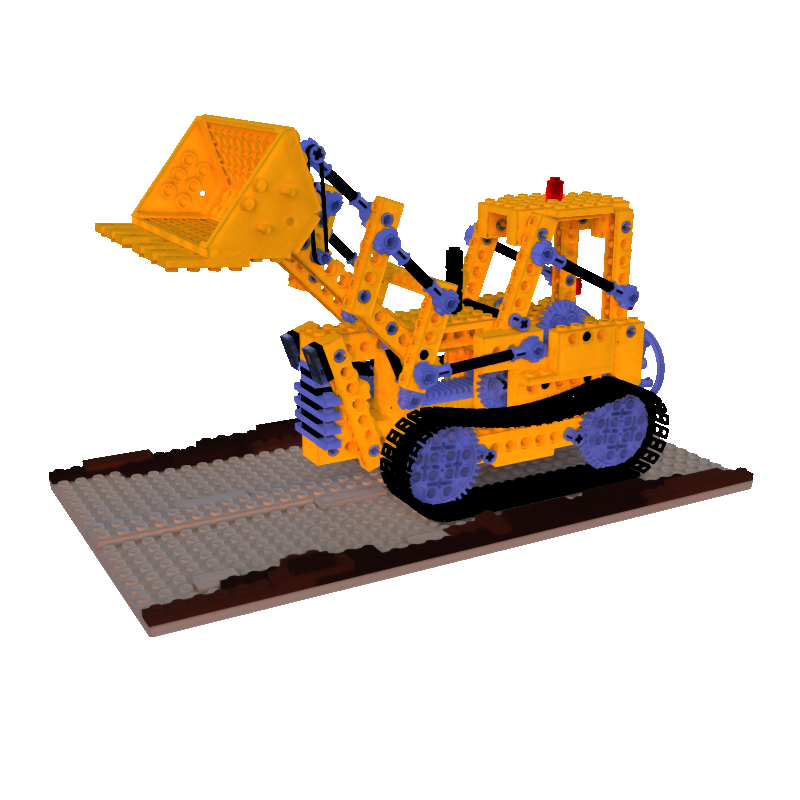}
    & \cellimg{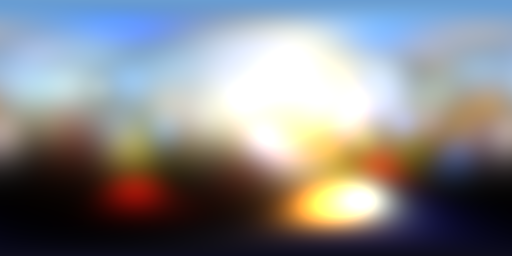}
    & \cellimg{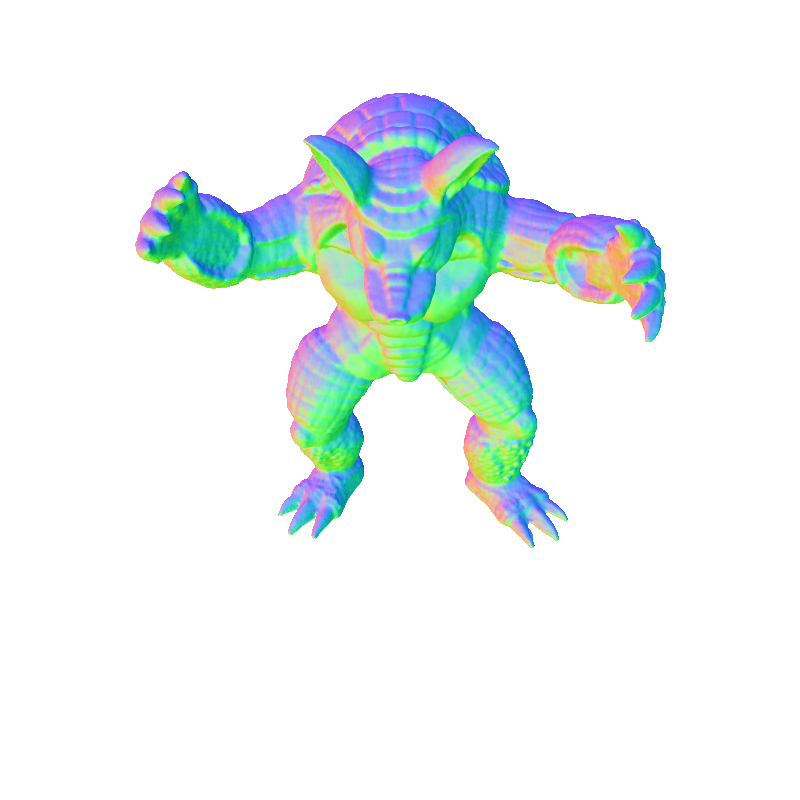}
    & \cellimg{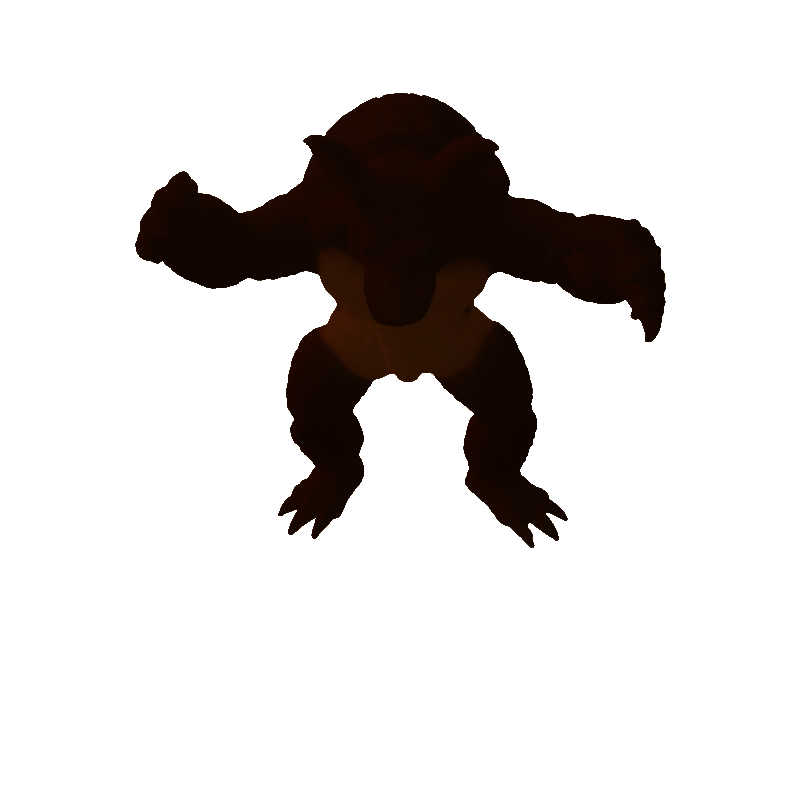}
    & \cellimg{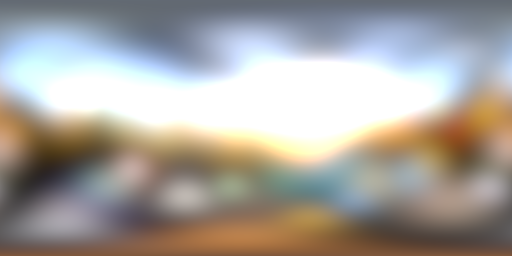}\\[-10pt]

    \textbf{GS-IR}
    & \cellimg{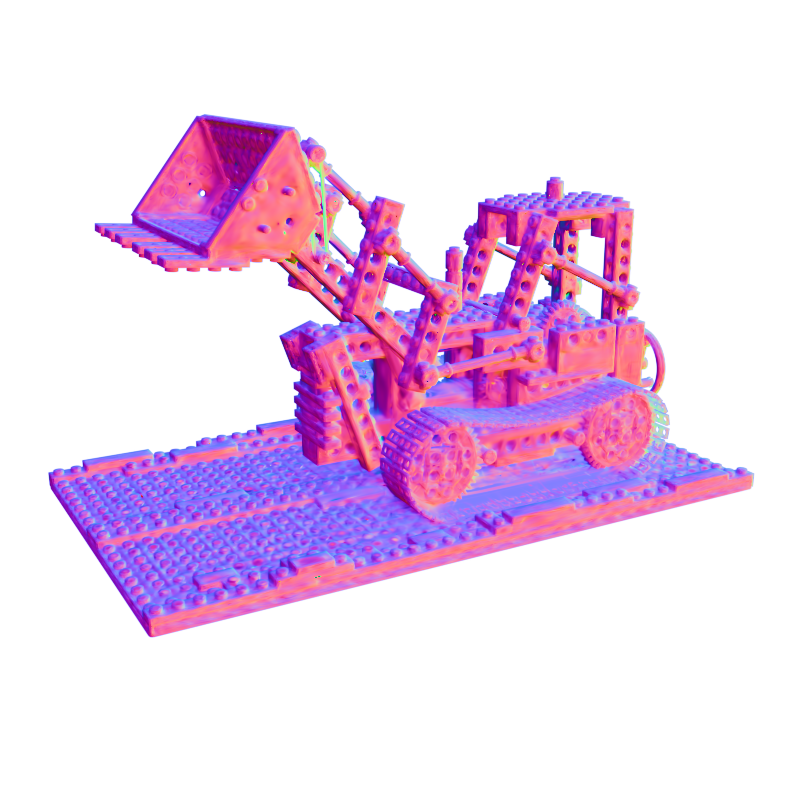}
    & \cellimg{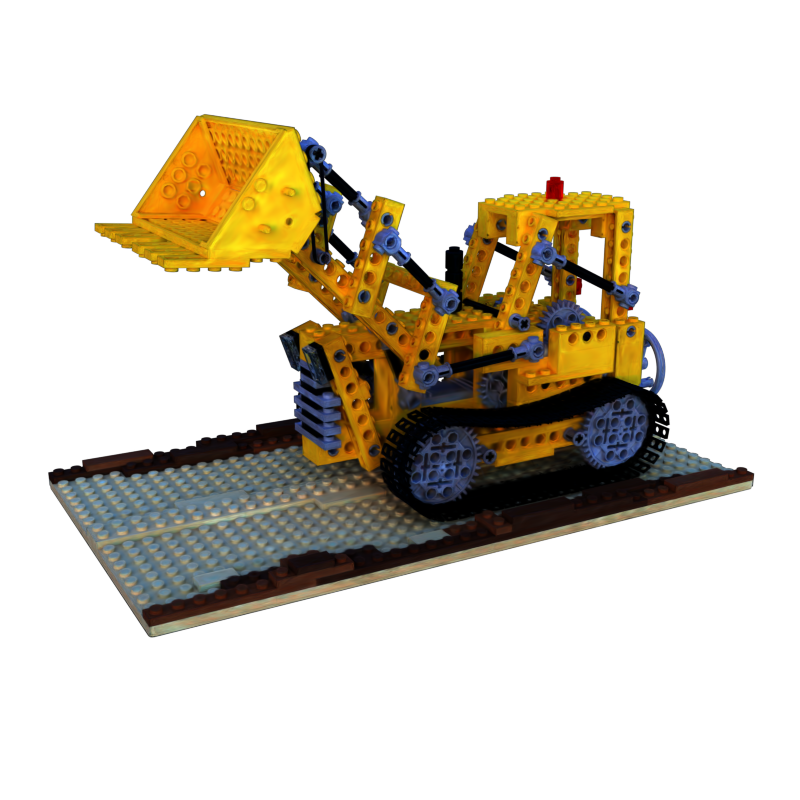}
    & \cellimg{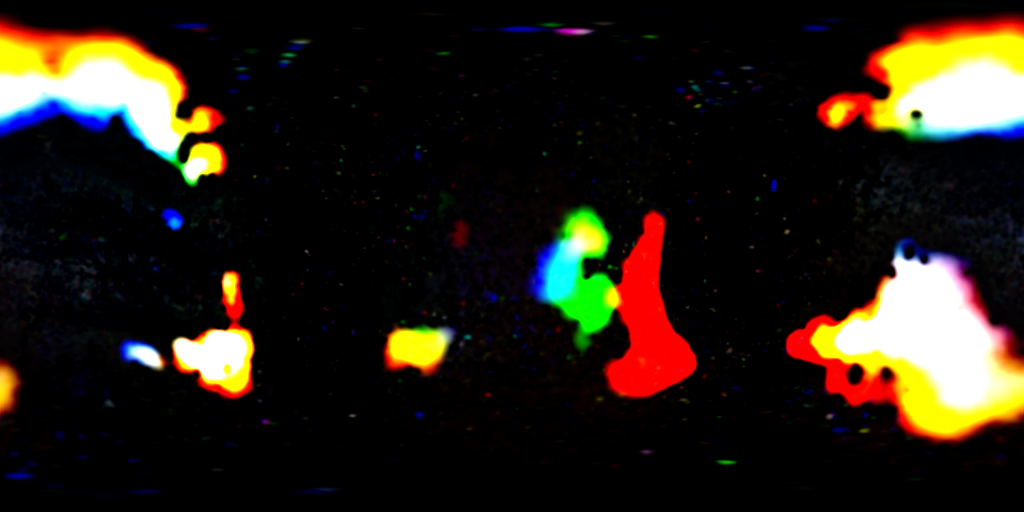}
    & \cellimg{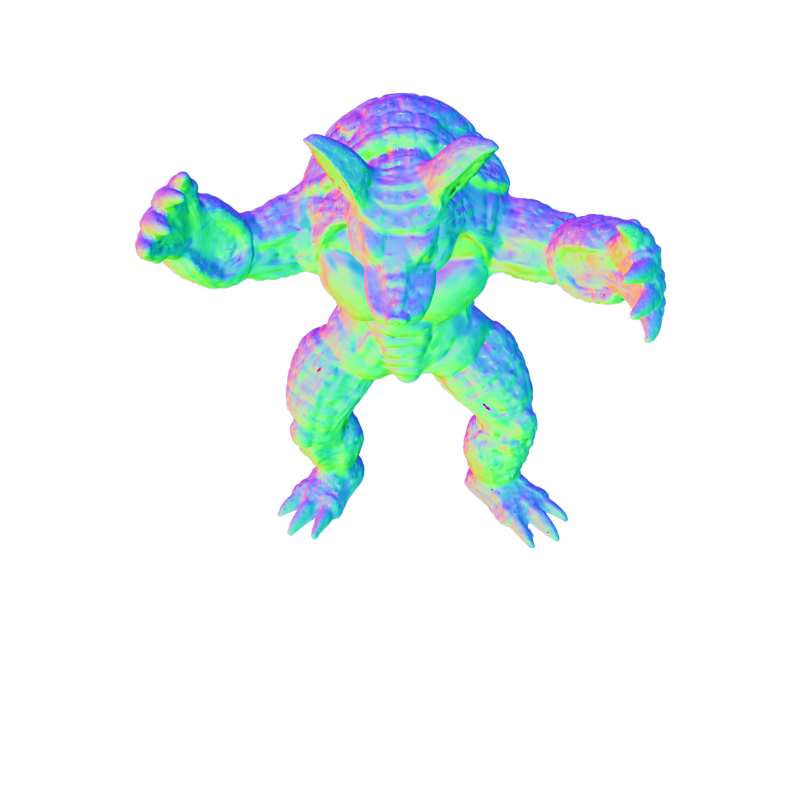}
    & \cellimg{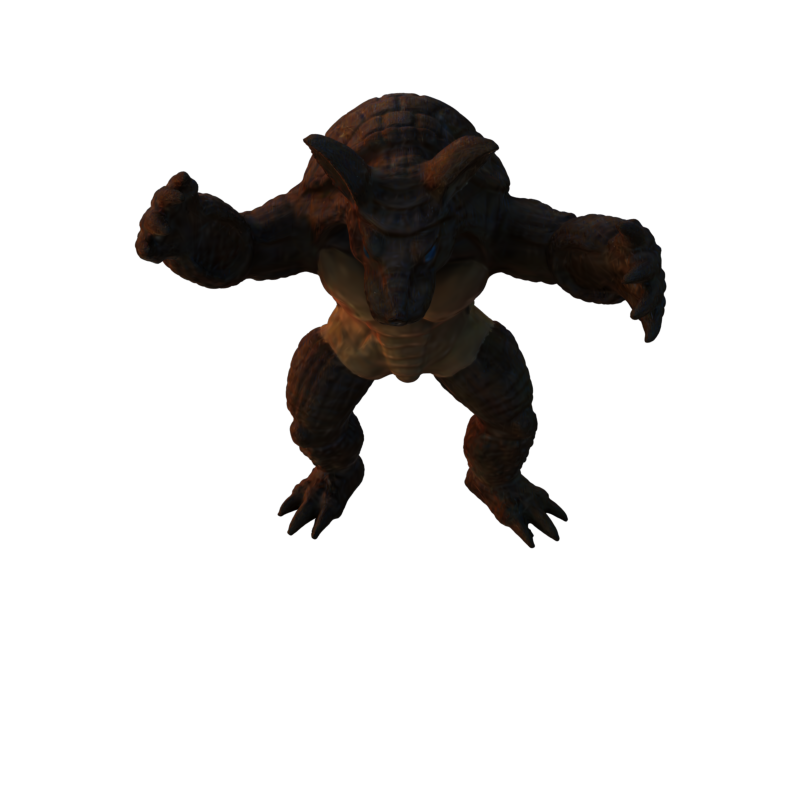}
    & \cellimg{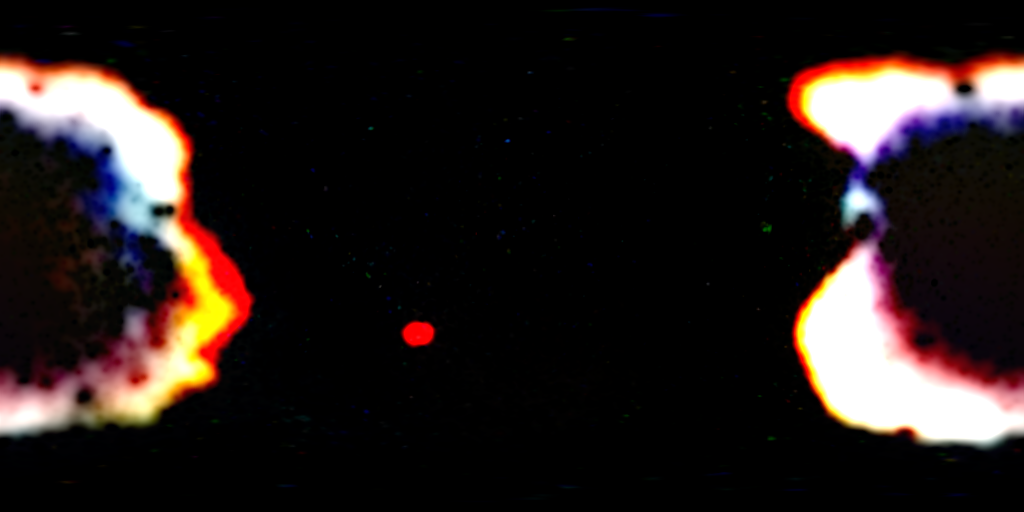}\\[-10pt]

    \textbf{IRGS}
    & \cellimg{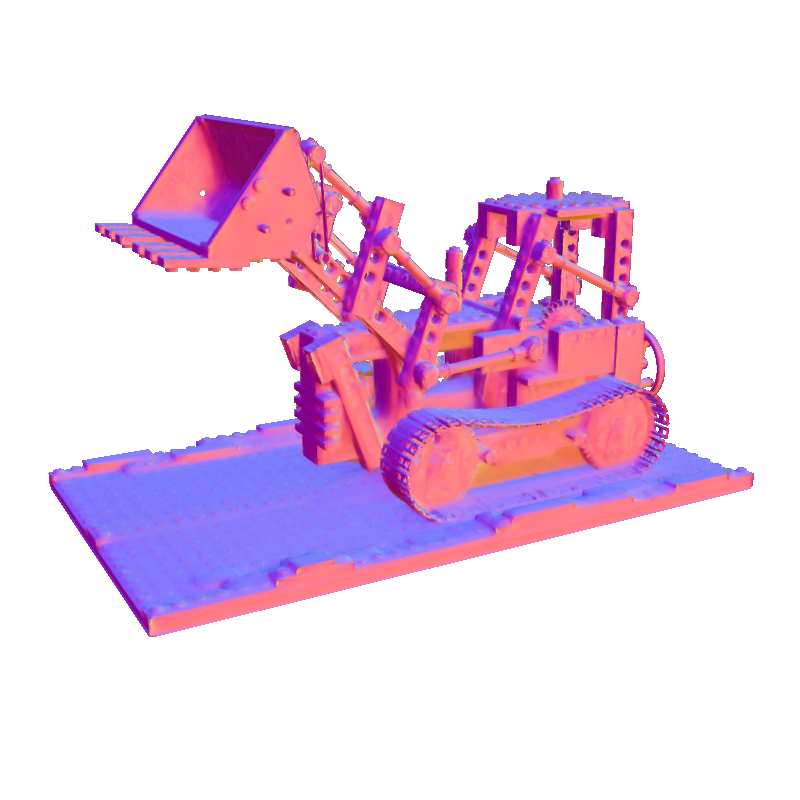}
    & \cellimg{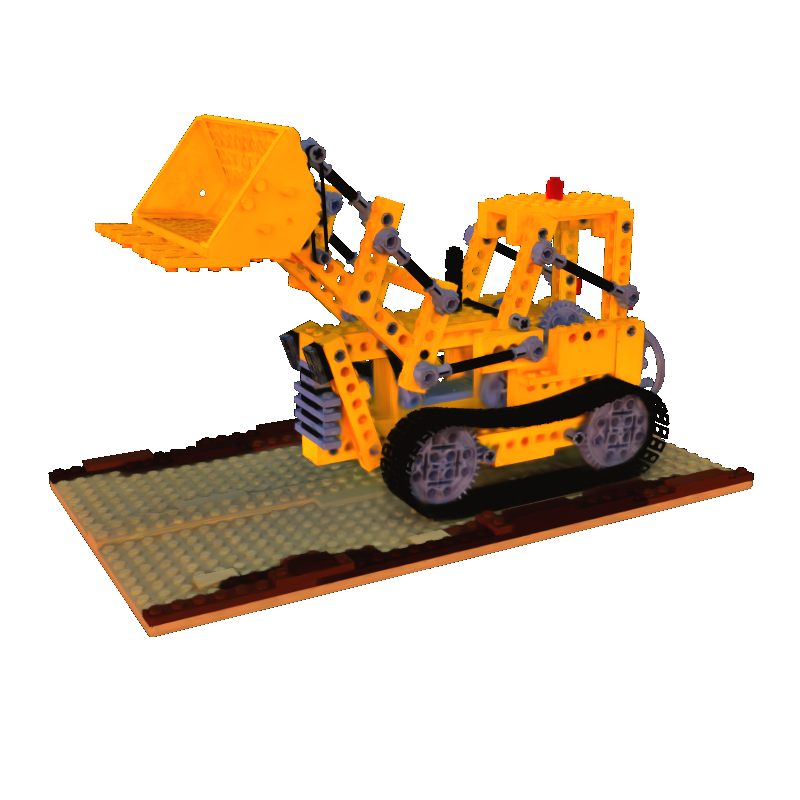}
    & \cellimg{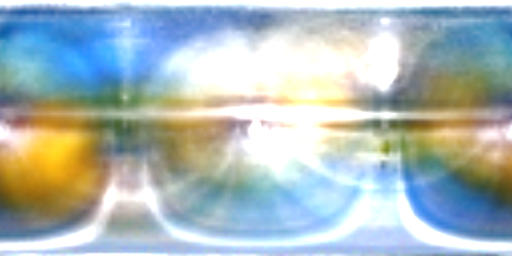}
    & \cellimg{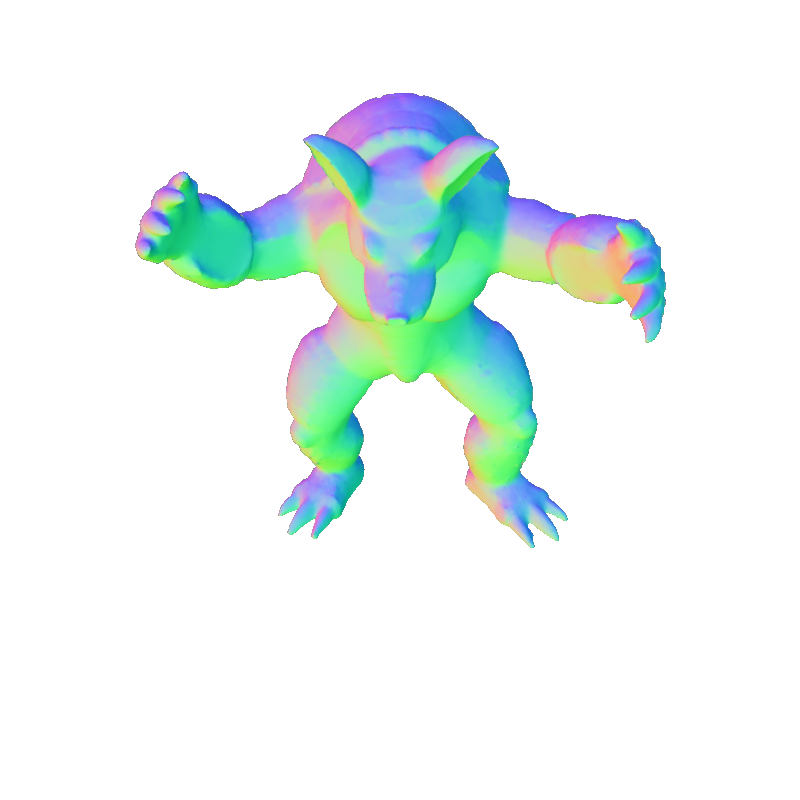}
    & \cellimg{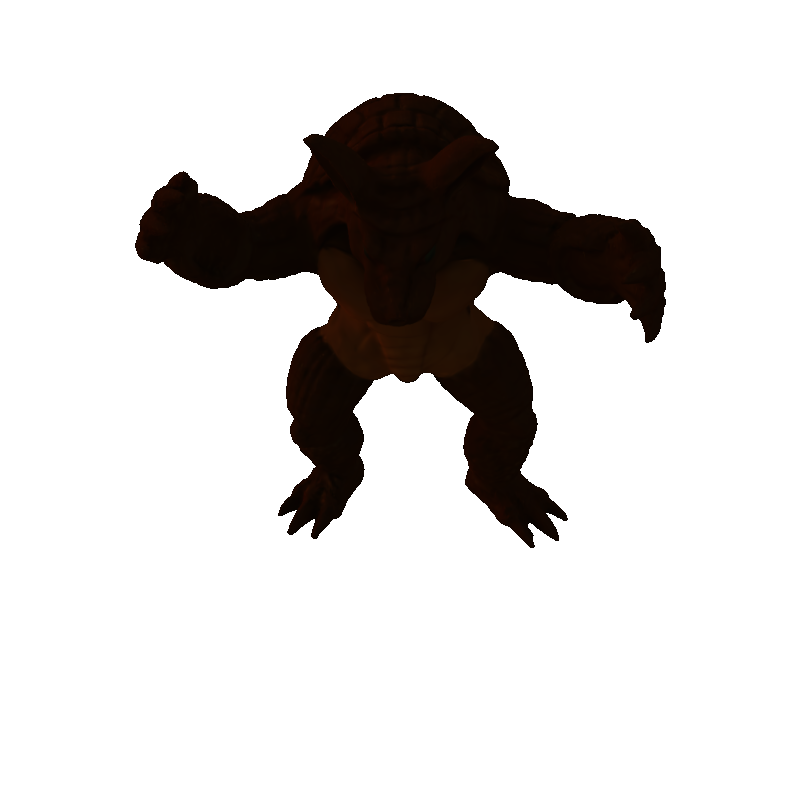}
    & \cellimg{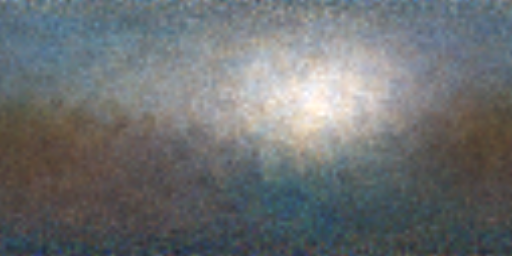}\\[-10pt]

    \textbf{Ours}
    & \cellimg{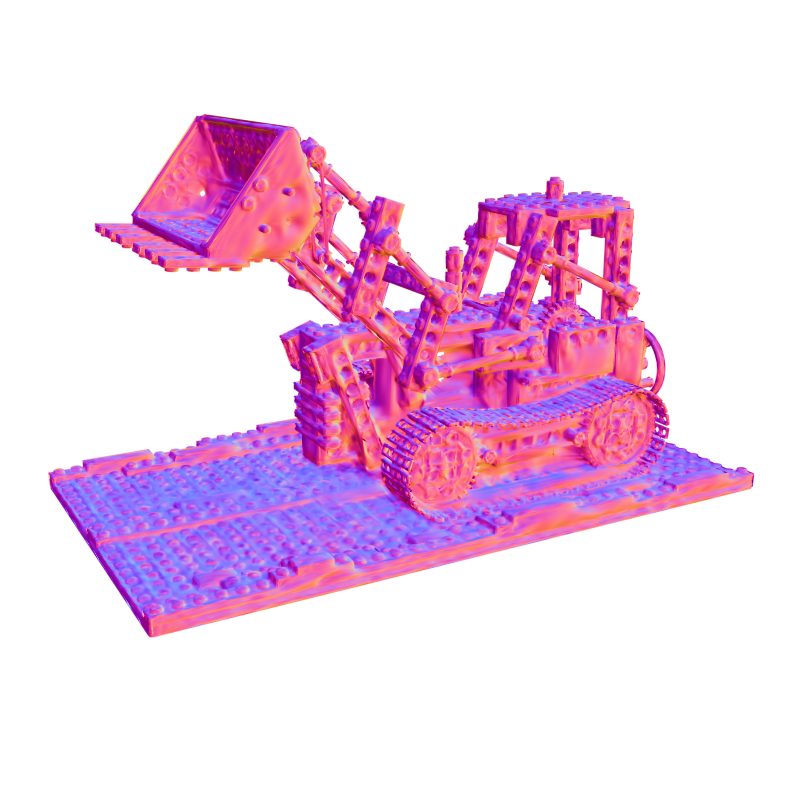}
    & \cellimg{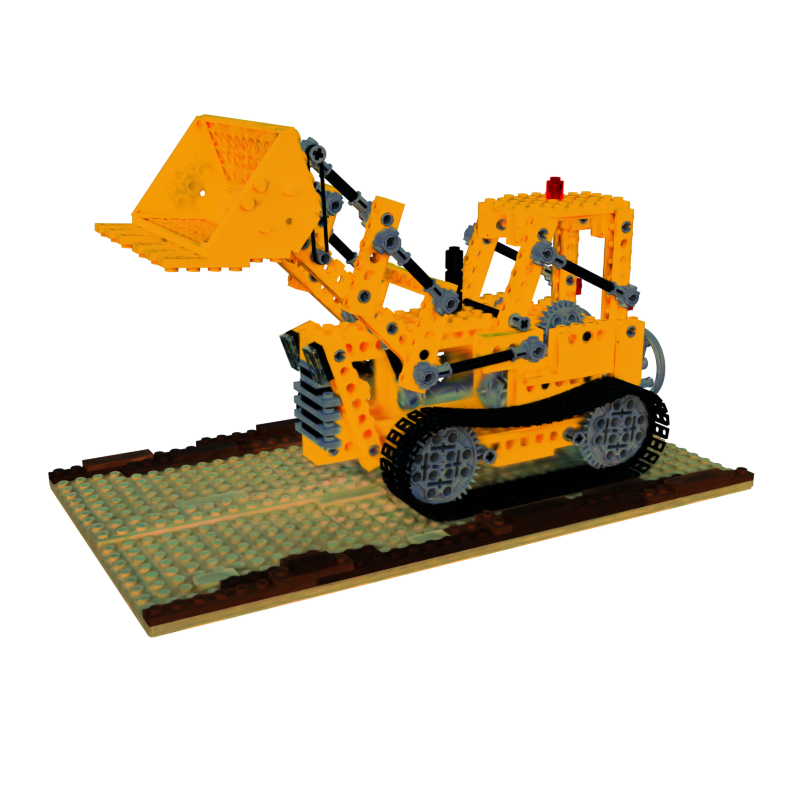}
    & \cellimg{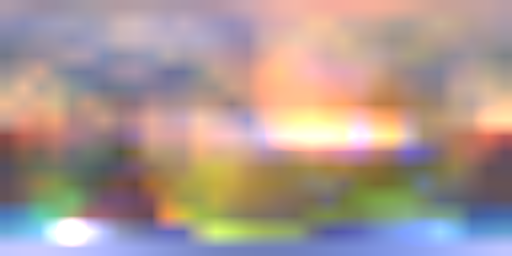}
    & \cellimg{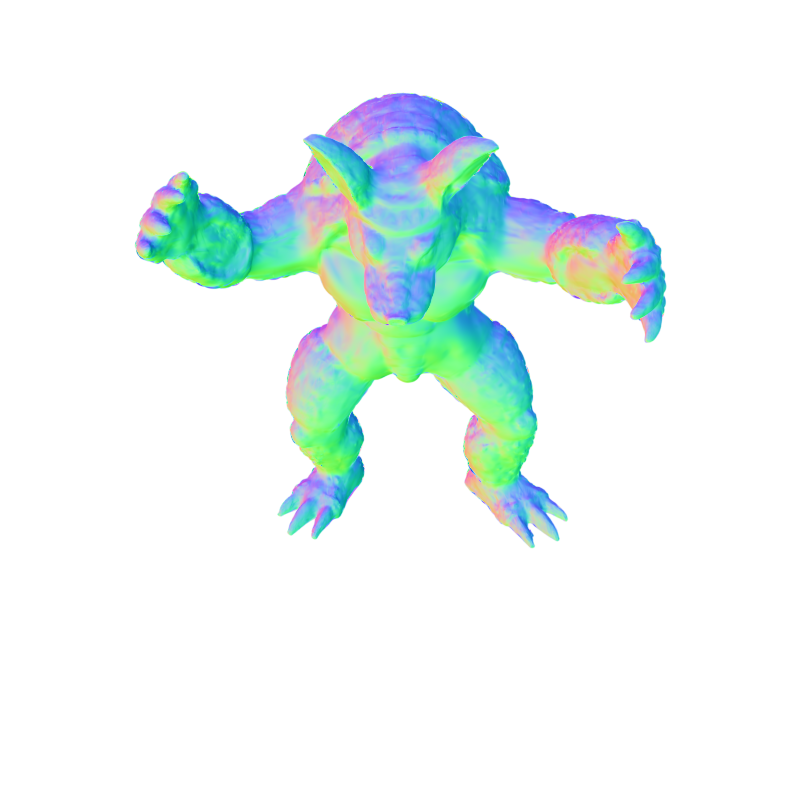}
    & \cellimg{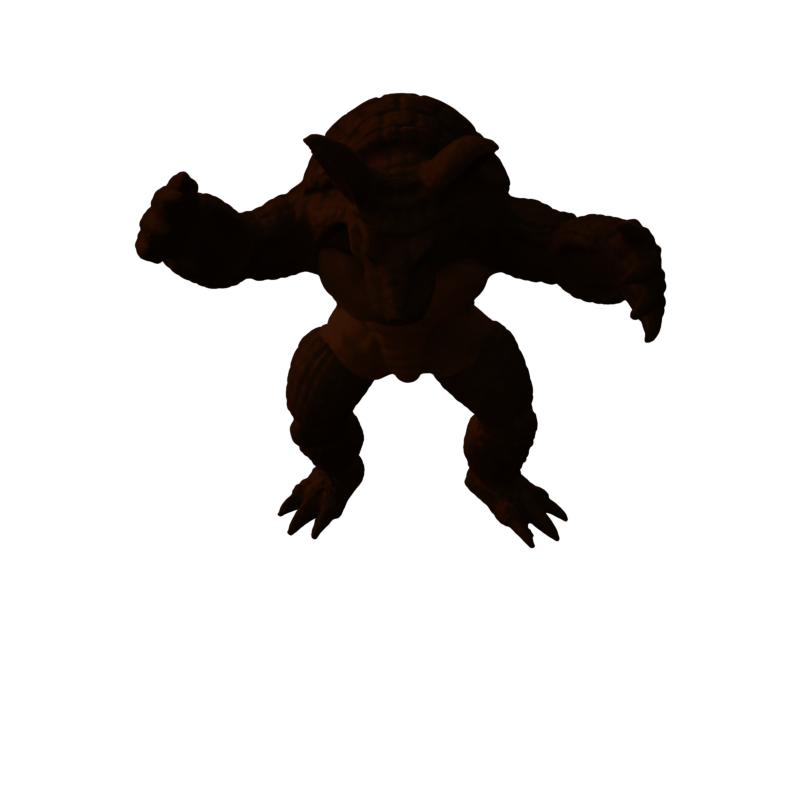}
    & \cellimg{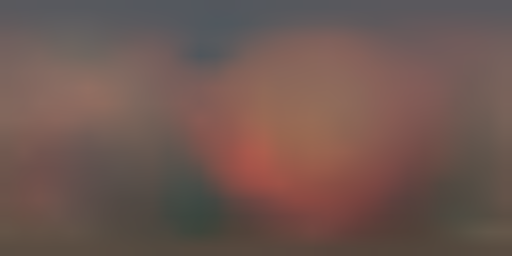}\\[-10pt]

    \textbf{GT}
    & \cellimg{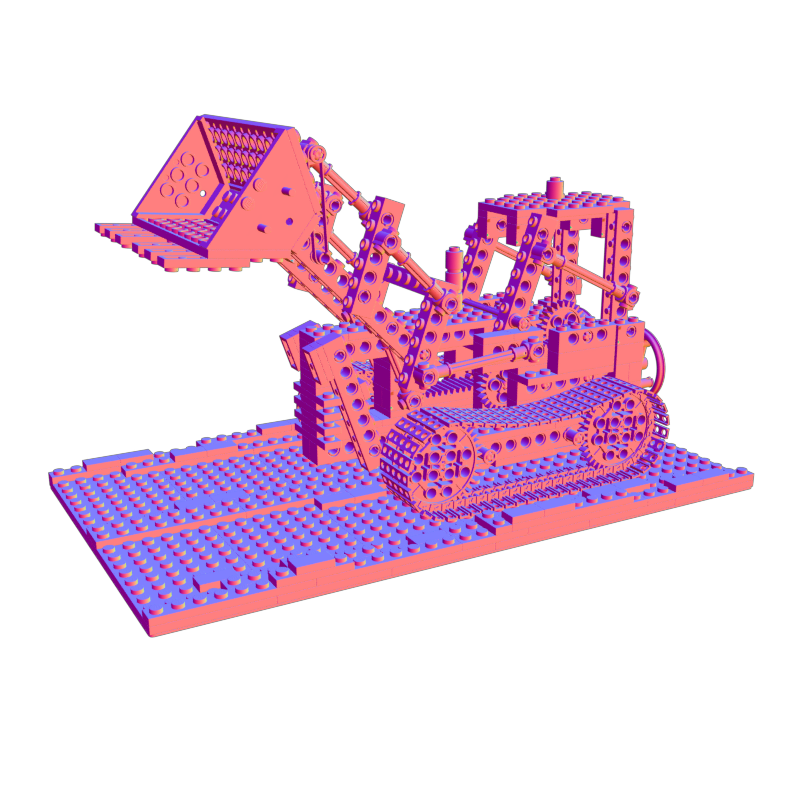}
    & \cellimg{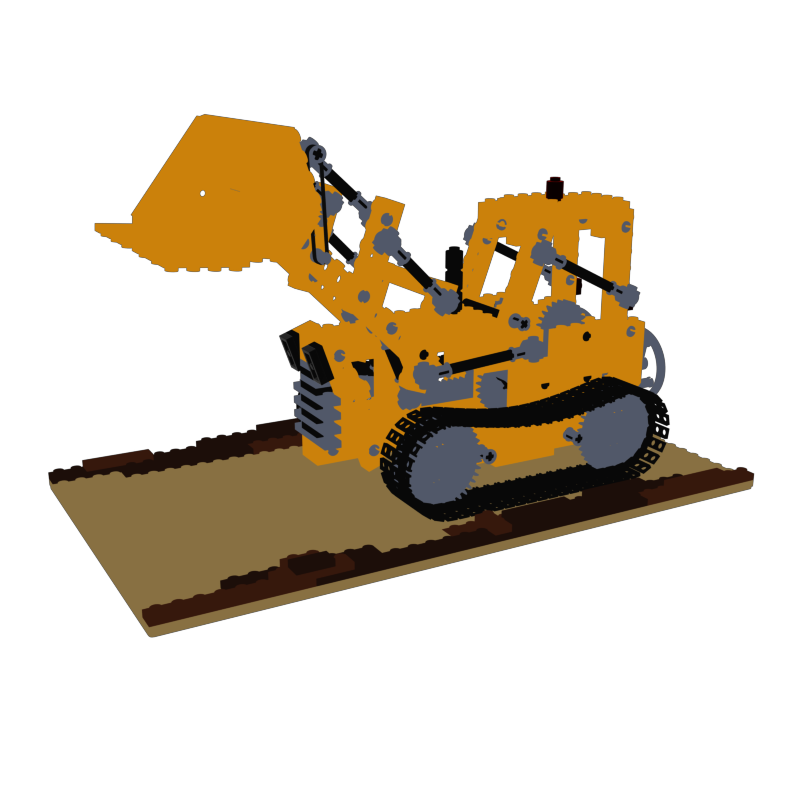} 
    & \cellimg{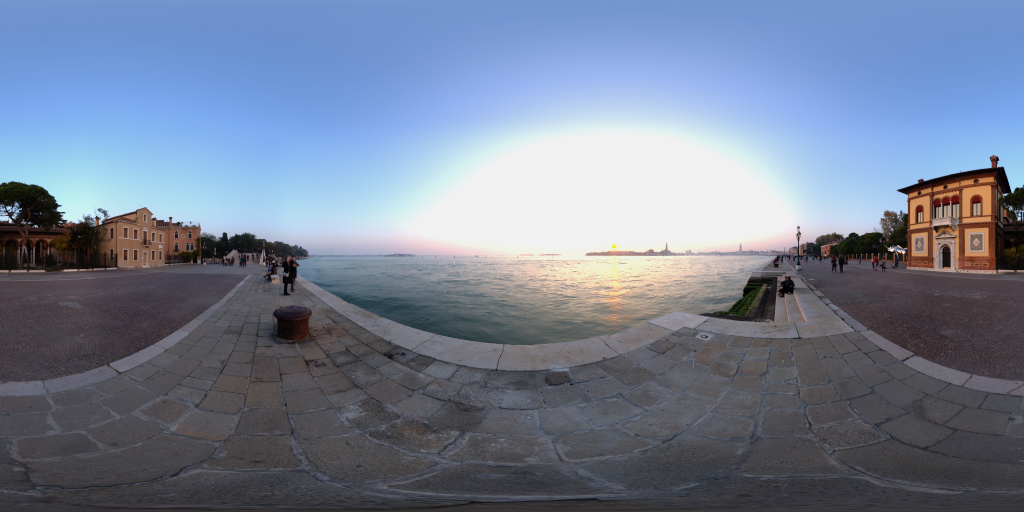}
    & \cellimg{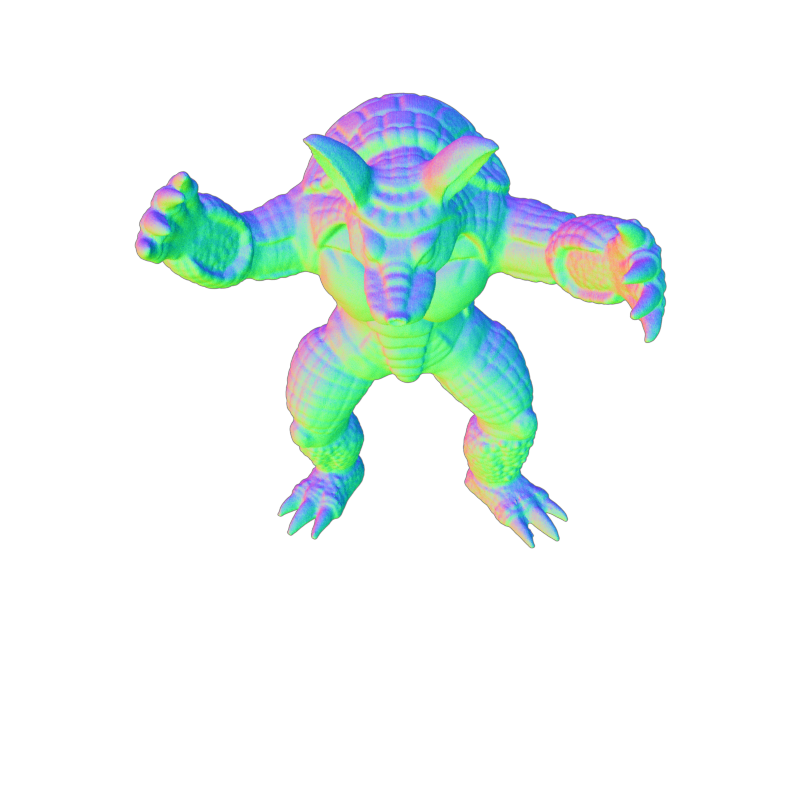}
    & \cellimg{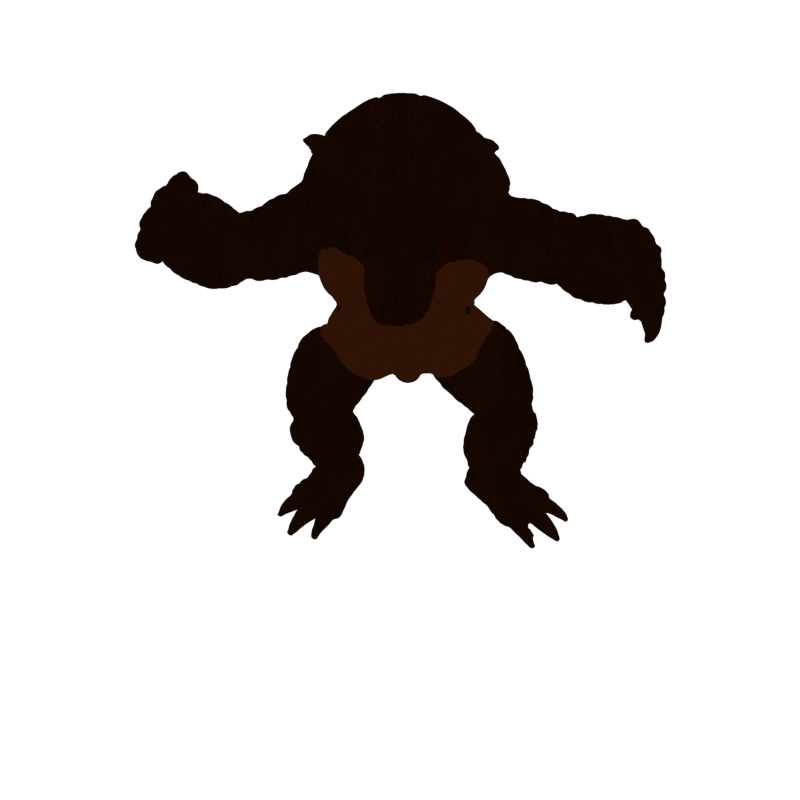}
    & \cellimg{figs/sunset.png}\\
\end{tabular}%
}

\caption{Qualitative comparison of albedo, normal and environmental lighting estimation on the TensoIR synthetic dataset \cite{jin:2023:tensoir}. }
\label{fig:comparison_grid}
\end{figure*}
\begin{figure}
    \centering
    \begin{subfigure}[t]{0.47\linewidth}
        \centering
        \includegraphics[width=\textwidth]{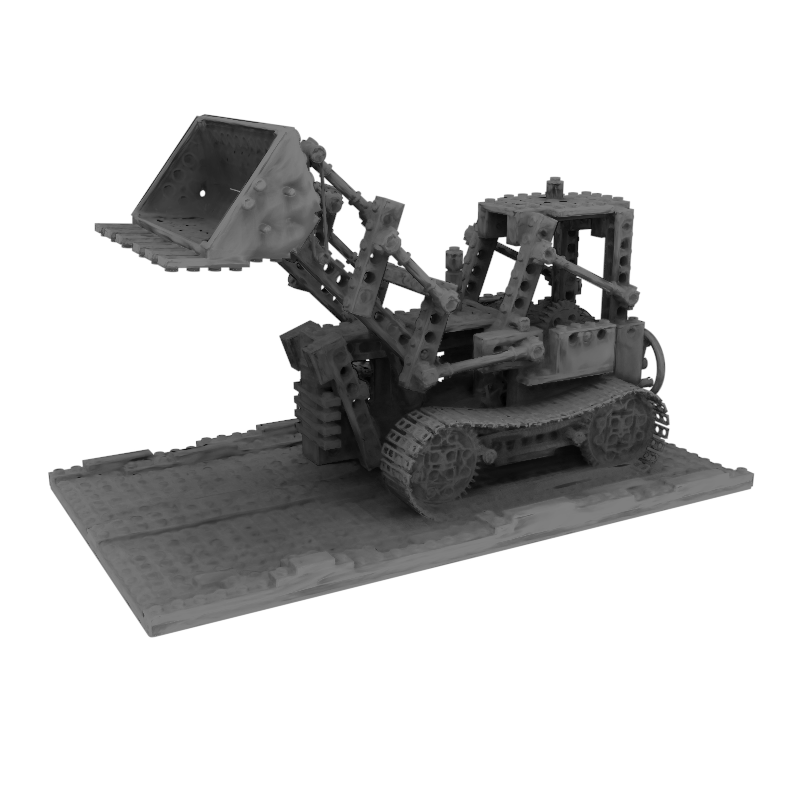}
        \caption{}
        \label{fig:occlusion_a}
    \end{subfigure}%
    \hfill
    \begin{subfigure}[t]{0.47\linewidth}
        \centering
        \includegraphics[width=\textwidth]{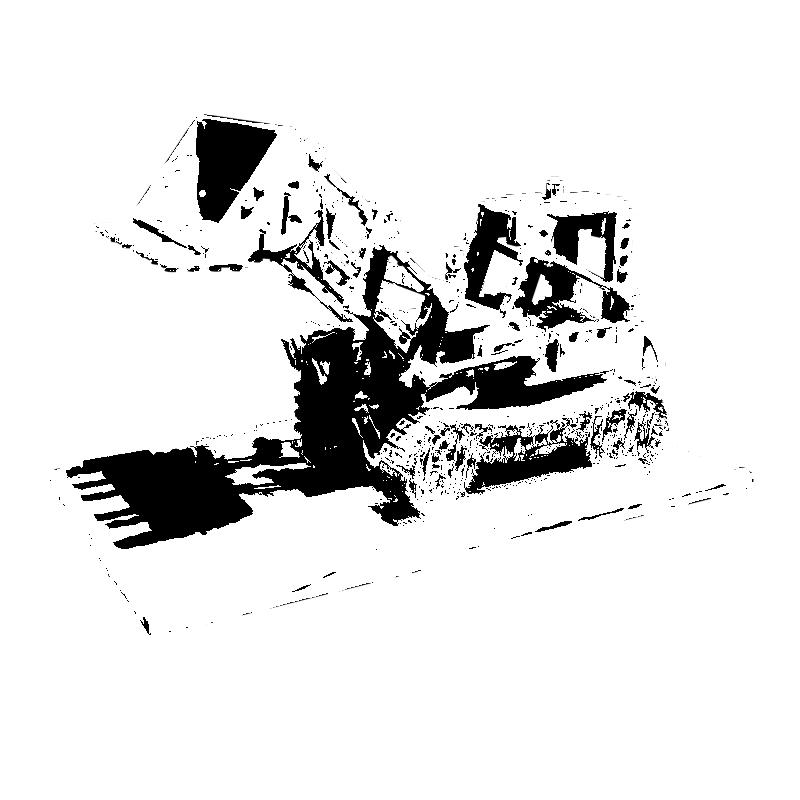}
        \caption{}
        \label{fig:occlusion_b}
    \end{subfigure}
    \vspace{4pt}
    \caption{(a) mean transmittance from all directions; (b) transmittance from a single direction.}
    \label{fig:occlusion}
\end{figure}

\textbf{Normal estimation:} Purely gradient-based normals derived from rendered depth are sensitive to noise and often lack multi-view consistency when applied to 3D Gaussian surfaces. We therefore assign each Gaussian $i$ a learnable unit normal $n_i$ and compute per-pixel predictions using splatting weights $w_i = T_i g_i \alpha_i$:
\begin{equation}
    \tilde{n}=\frac{\sum^N_{i=1}w_i n_i}{\sum^N_{i=1}w_i+\epsilon_n}, \quad \hat{n}=\frac{\tilde{n}}{\|\tilde{n}\|_2+\epsilon_n},
\end{equation}
where $\epsilon_n$ is a small constant for numerical stability. As a geometry-anchored target, we derive pseudo-normals from expected depth $z_E$. World-space normals $\hat{n}_\nabla$ are obtained from image-space gradients $\nabla z_E=(\partial_x z_E,\partial_y z_E)$, and predictions are aligned to this estimate with
\begin{equation}
    \mathcal{L}_{n-p}=||\hat{n}-\hat{n}_\nabla||.
\end{equation}
To further smooth predictions, we follow GS-IR~\cite{liang:2024:gs-ir} and add a total variation (TV) loss $TV_{normal}$.

The overall normal objective is
\begin{equation}
    \mathcal{L}_n = \mathcal{L}_{n-p} + \lambda_{n-TV}TV_{normal},
\end{equation}
and is optimized jointly with the photometric loss $\mathcal{L}_c$ of the splatting model.

\begin{figure*}[t]
  \centering
  \setlength{\tabcolsep}{2pt}
  \renewcommand{\arraystretch}{0}

  \begin{subfigure}[t]{0.24\textwidth}
    \centering
    \begin{tabular}{@{}cc@{}}
      \includegraphics[width=0.48\linewidth]{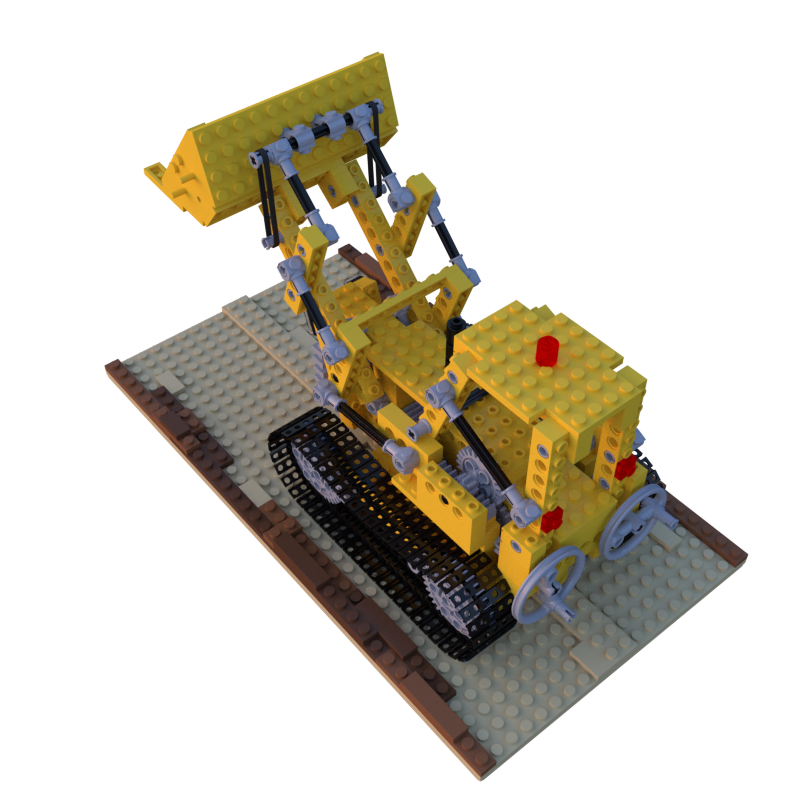} &
      \includegraphics[width=0.48\linewidth]{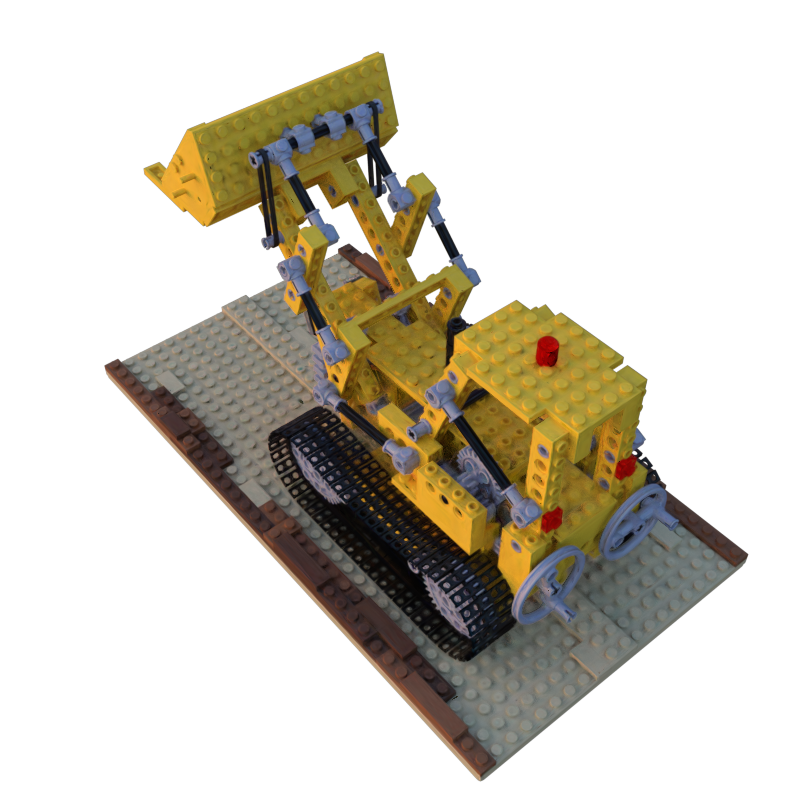} \\
      \includegraphics[width=0.48\linewidth]{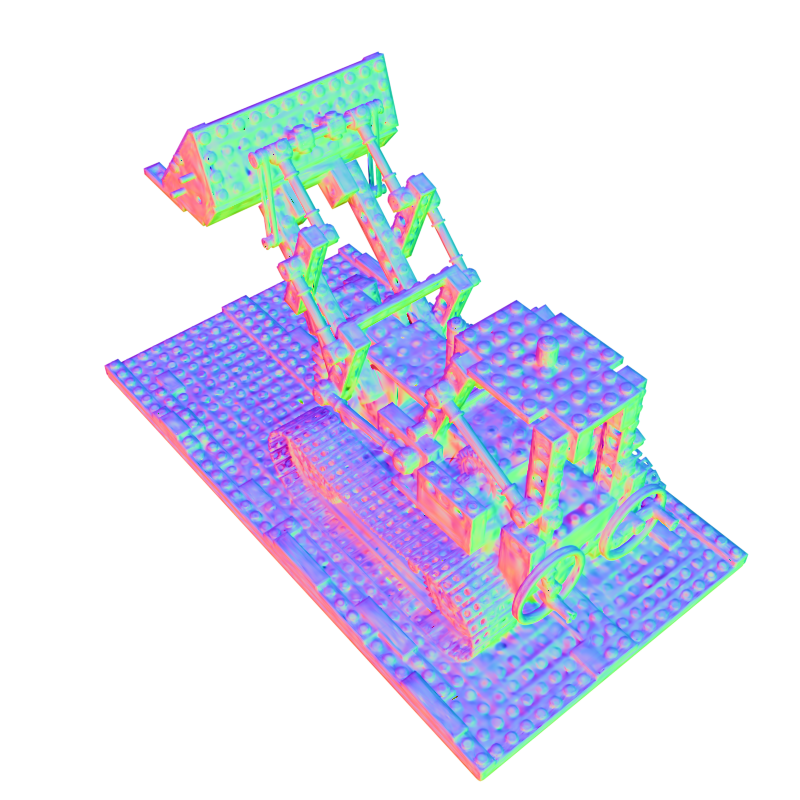} &
      \includegraphics[width=0.48\linewidth]{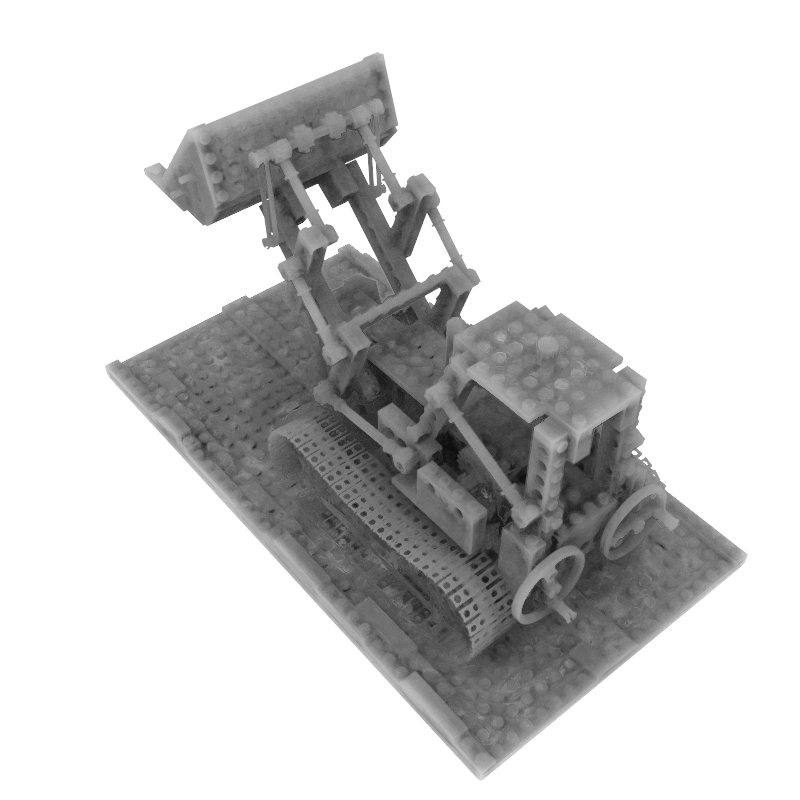}
    \end{tabular}
    \caption{Lego}
    \label{fig:blockA}
  \end{subfigure}\hfill
  \begin{subfigure}[t]{0.24\textwidth}
    \centering
    \begin{tabular}{@{}cc@{}}
      \includegraphics[width=0.48\linewidth]{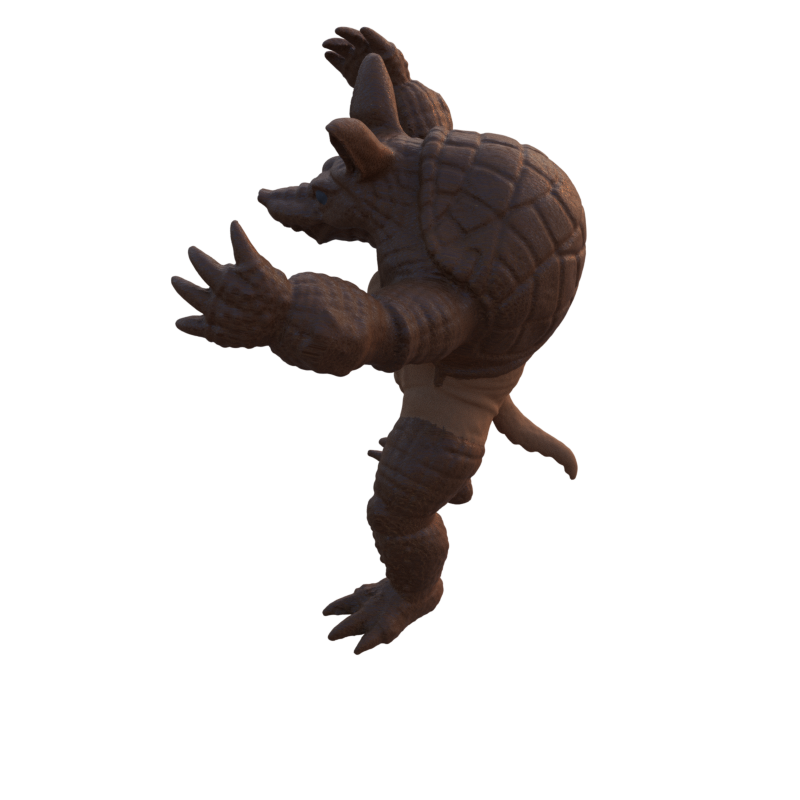} &
      \includegraphics[width=0.48\linewidth]{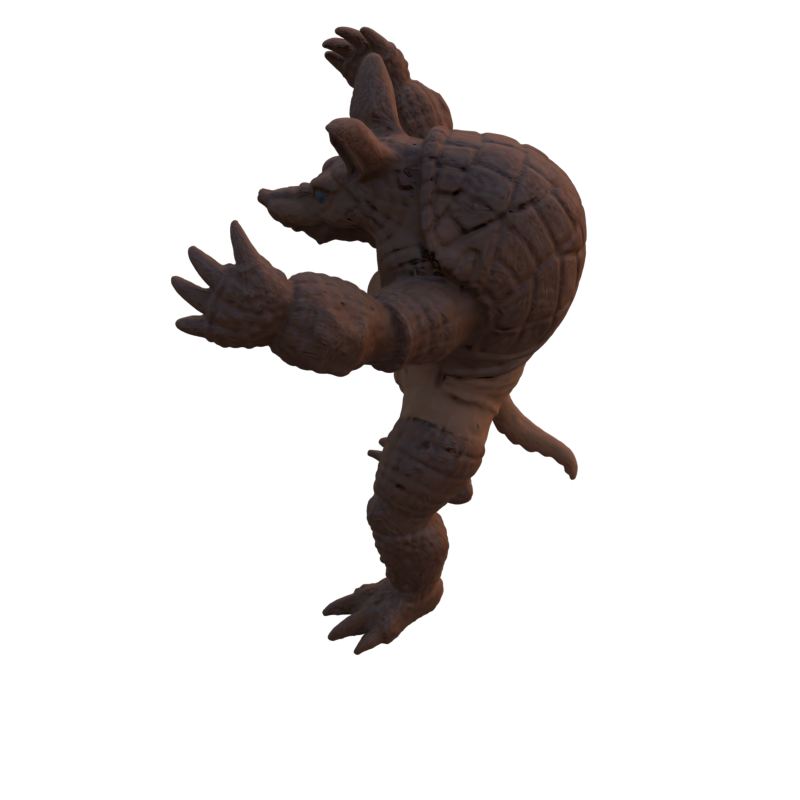} \\
      \includegraphics[width=0.48\linewidth]{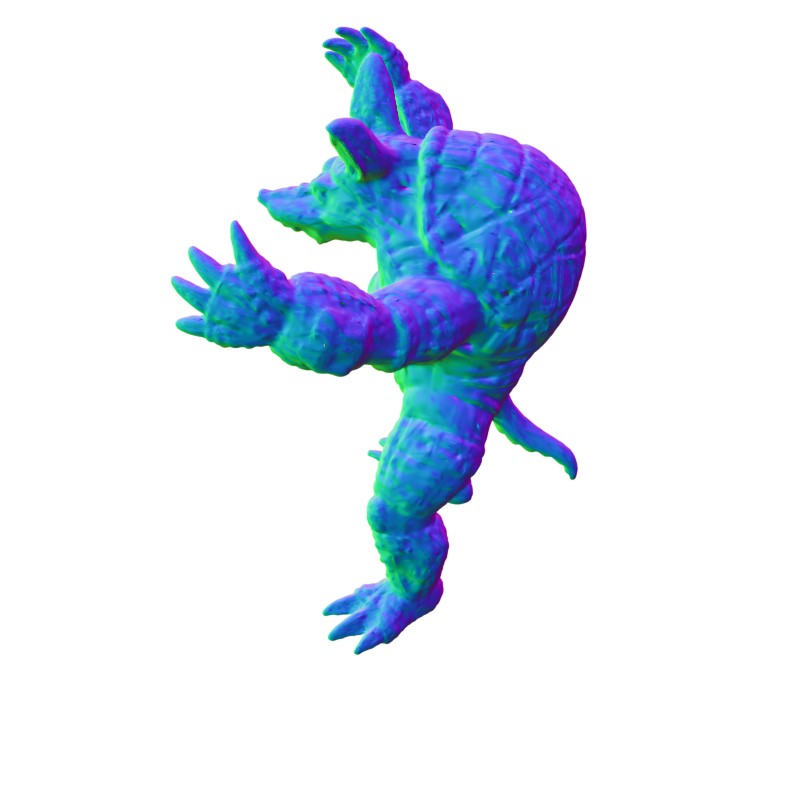} &
      \includegraphics[width=0.48\linewidth]{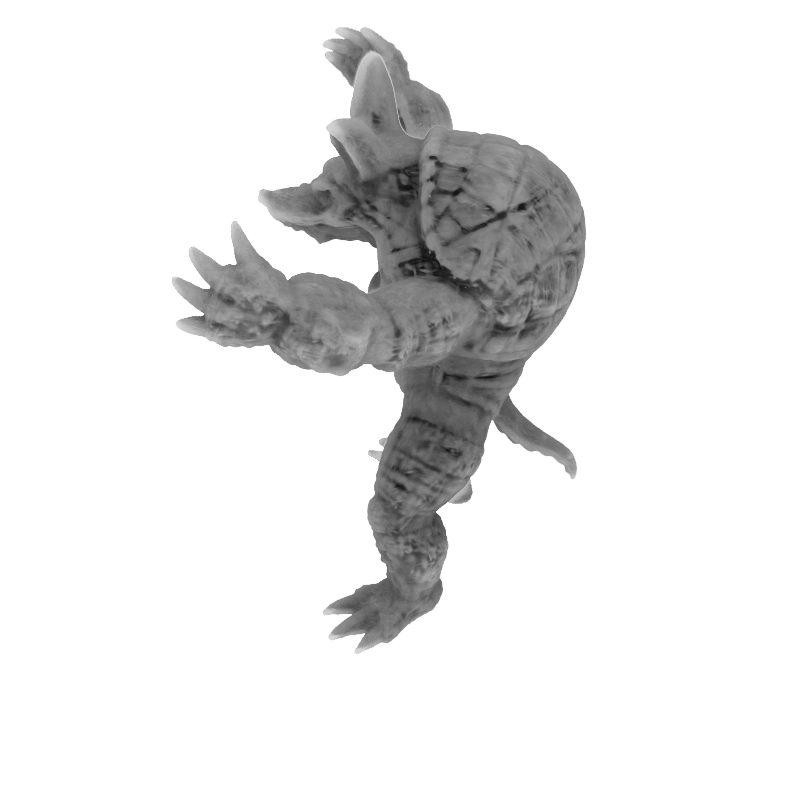}
    \end{tabular}
    \caption{Armadillo}
    \label{fig:blockB}
  \end{subfigure}\hfill
  \begin{subfigure}[t]{0.24\textwidth}
    \centering
    \begin{tabular}{cc}
      \includegraphics[width=0.48\linewidth]{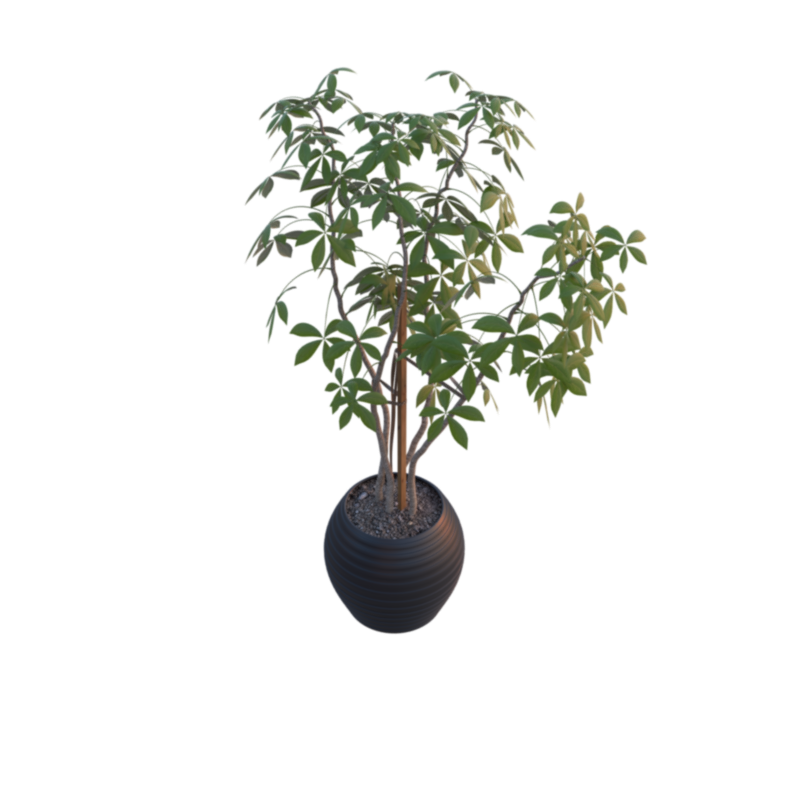} &
      \includegraphics[width=0.48\linewidth]{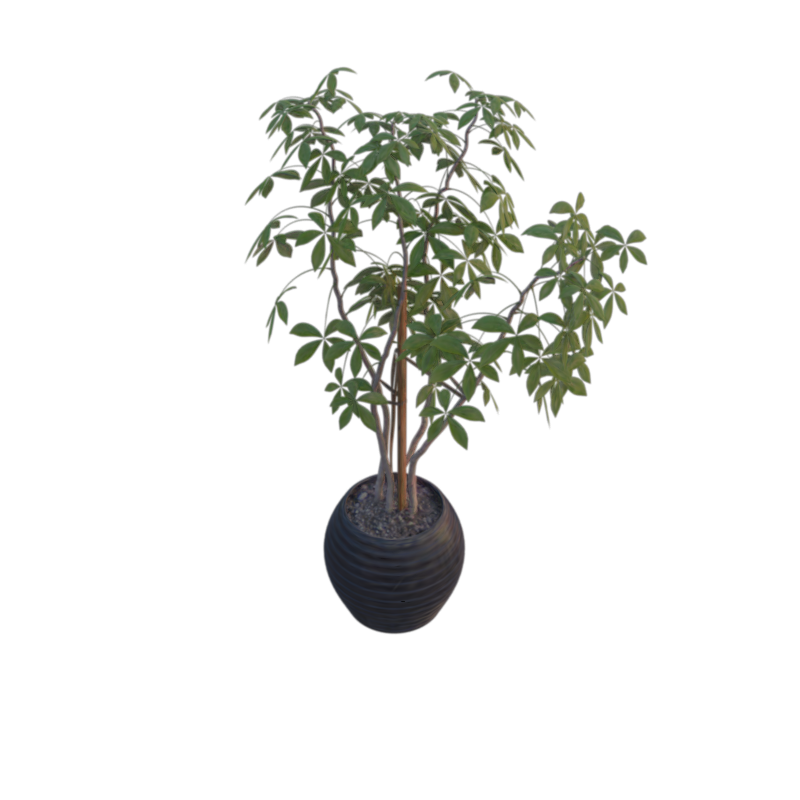} \\
      \includegraphics[width=0.48\linewidth]{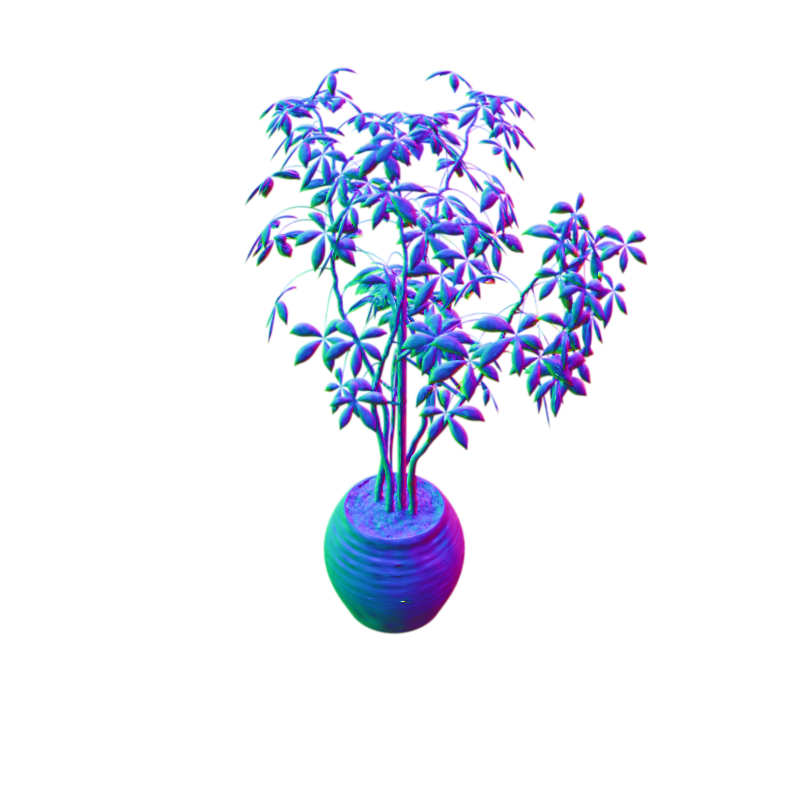} &
      \includegraphics[width=0.48\linewidth]{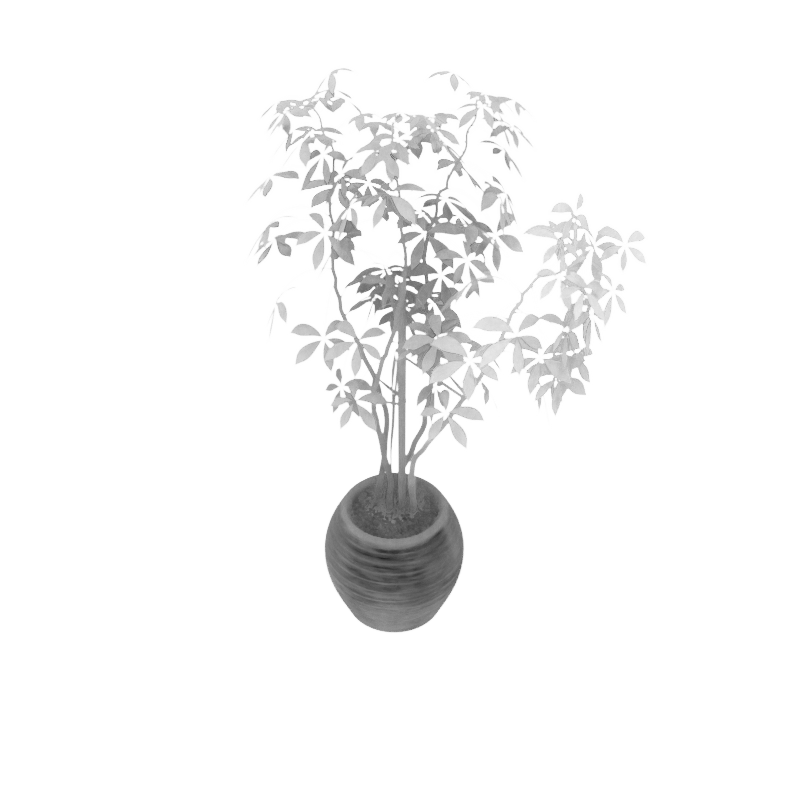}
    \end{tabular}
    \caption{Ficus}
    \label{fig:blockC}
  \end{subfigure}
  \begin{subfigure}[t]{0.24\textwidth}
    \centering
    \begin{tabular}{cc}
      \includegraphics[width=0.48\linewidth]{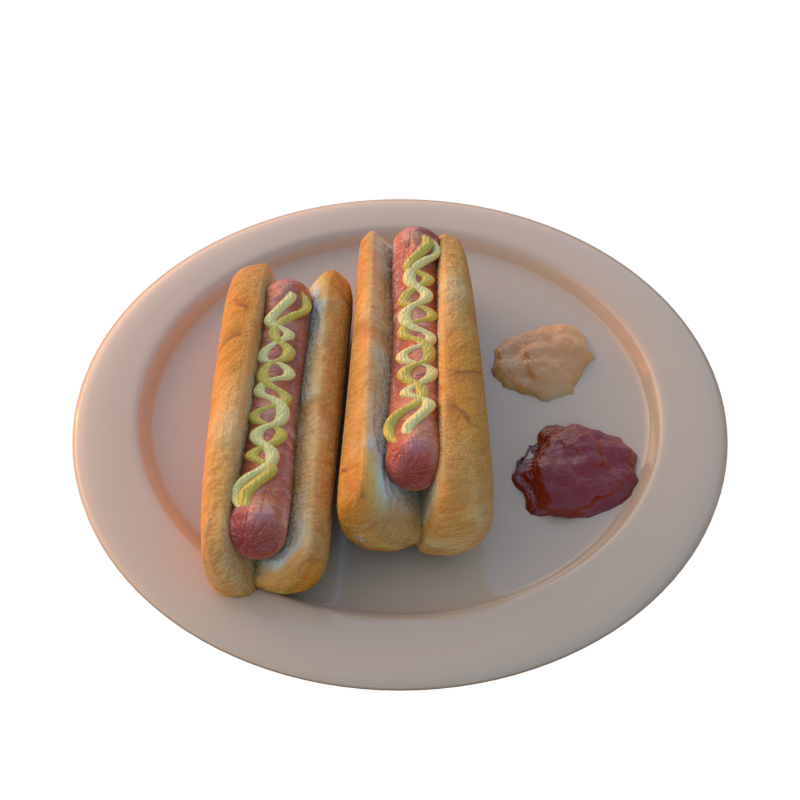} &
      \includegraphics[width=0.48\linewidth]{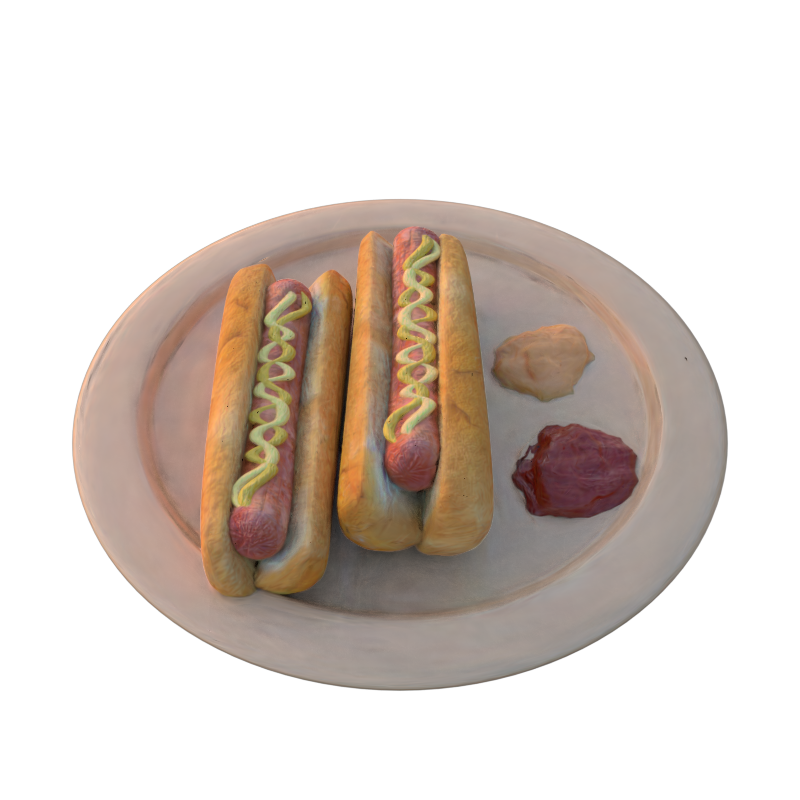} \\
      \includegraphics[width=0.48\linewidth]{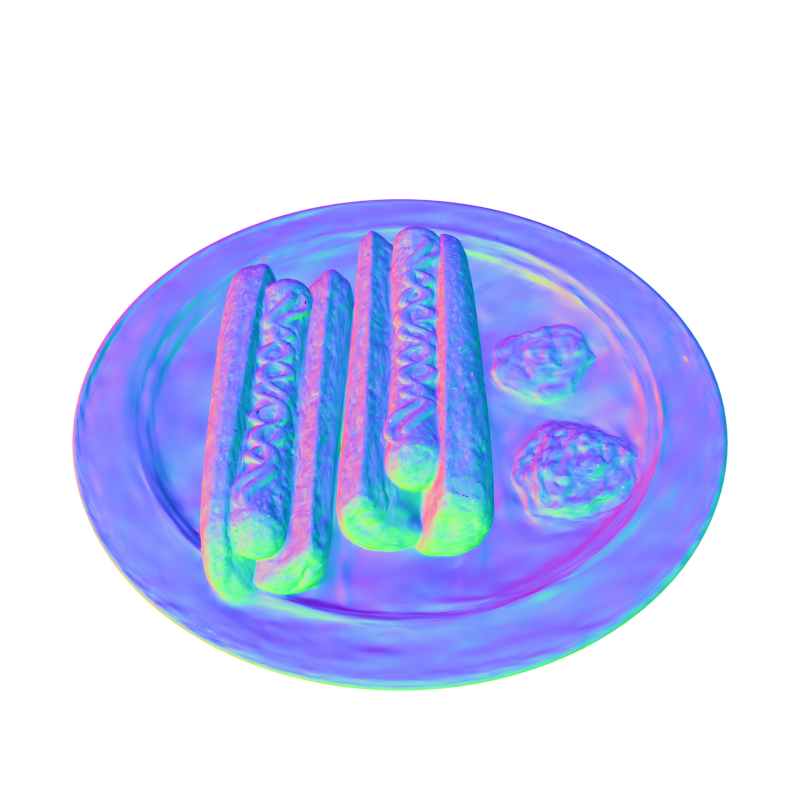} &
      \includegraphics[width=0.48\linewidth]{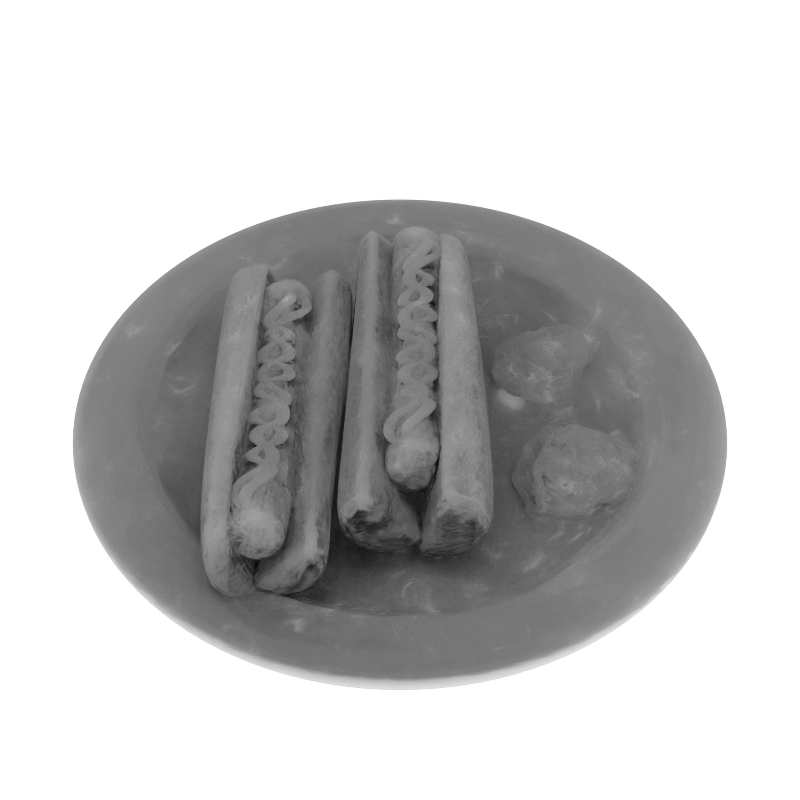}
    \end{tabular}
    \caption{Hotdog}
    \label{fig:blockD}
  \end{subfigure}

  \vspace{4pt}

  \caption{Visual results on the TensoIR synthetic dataset \cite{jin:2023:tensoir}. Each model is displayed in order from left to right and from top to bottom: ground truth, novel-view synthesis (NVS), normal map, and mean visibility.}
  \label{fig:all_models}
\end{figure*}

\subsection{Point-Based Ray Tracing}
In our pipeline, the ray tracer is not used as a replacement renderer. It serves as a sparse visibility oracle for evaluating transmittance along sampled light directions.

The 3DGS rasterizer is a highly optimized, tile-based renderer: for each screen-space tile it collects all projected splats overlapping the tile, sorts them in front-to-back order, and evaluates color and opacity via elliptical weighted average (EWA) splatting \cite{zwicker:2002:ewa} in that order. This design delivers excellent throughput for primary views but is tied to camera projection and ordered blending per tile.

A point-based ray tracer allows us to evaluate Gaussian splatting from any point in space in any direction. The 3D starting points for our rays are computed from the expected-depth map $z_E$. Following 3D Gaussian Ray Tracing (3DGRT) \cite{loccoz:2024:3dgrt}, each 3D Gaussian is enclosed by an anisotropically stretched regular icosahedron and the triangles of this proxy mesh are inserted into a BVH for fast ray-triangle traversal. The proxy is scaled so that it tightly bounds the region where the particle's response exceeds a small minimum $\epsilon$. The scale explicitly incorporates the particle's opacity $\alpha$ so that nearly transparent splats receive proportionally tighter bounds (''adaptive clamping''), which reduces false-positive hits during traversal. Rays then query the BVH to gather the next batch of intersections, evaluate contributions in sorted order, and early-terminate once the accumulated transmittance reaches a preset threshold.

For occlusion-only queries, we use the fact that transmittance along a ray can be accumulated multiplicatively $ T \leftarrow T \cdot (1-\Delta_{a_i})$ where $\Delta_{a_i}$ is the per-hit attenuation computed from the Gaussian's line integral along the ray segment covered by the proxy. Because multiplication commutes, the hits need not be depth-sorted for this visibility pass. They can be processed in arbitrary order while maintaining correctness for $T$. We early-terminate when $T\leq T_{stop}$. In practice, $T_{stop}$ should match the early-out used by the rasterizer to keep consistency between visibility estimated by ray tracing and the screen-space renderer. To make the ray tracer a drop-in companion to standard 3DGS (so we can reuse pretrained point clouds and the existing rasterizer), we evaluate Gaussians on a ray with the same EWA-splatting formulation as the rasterizer. For a ray $r(t)=o+td$, we instantiate a one-pixel virtual camera at $o$ whose optical axis is $d$ (i.e., a local camera frame aligned with the ray). We then reuse the unmodified rasterizer to splat exactly the Gaussians reported by the BVH for that ray into the center pixel of this virtual camera. This yields consistent blending with the original tile-based renderer while allowing us to drive the evaluation from rays rather than from a full image plane.

Alternative ray-based renderers make different evaluation choices that deviate from EWA splatting: 3DGRT \cite{loccoz:2024:3dgrt} evaluates each Gaussian at the maximum response along the ray and processes batches of intersections in sorted order using a k-buffer scheme. IRGS \cite{gu:2025:irgs} collapses 3D Gaussians to planar 2D Gaussian disks \cite{huang:2024:2dgs} to obtain a closed-form ray-splat intersection. These choices are effective but generally require training with their respective renderer to avoid a mismatch with classic 3DGS splatting.

\subsection{Optimization}
Our optimization method consists of a two-stage strategy, as shown in Fig. \ref{fig:overview}.

\textbf{Initial stage.} The first step in our optimization pipeline is pretraining a standard Gaussian model with the extension that we jointly train the normal $\hat{n}$.
During this stage, we observe partially transparent regions where background color can leak through the object silhouette. To address this, we add an opacity-mask loss
\begin{equation}
    \mathcal{L}_{mask}=||\hat{\alpha}-M||_1,
\end{equation}
where $\hat{\alpha}$ is the rendered opacity and $M$ is the binary foreground mask. By measuring the deviation of Gaussian splats from the expected surface, we find that this term reduces that deviation and pulls splats closer to the surface, yielding a better-defined geometry proxy for our ray-tracing pass.

We use the color reconstruction loss $\mathcal{L}_c$ from standard Gaussian splatting, the normal loss $\mathcal{L}_n$, and the opacity-mask loss $\mathcal{L}_{mask}$. The full loss function for our initial stage is:
\begin{equation}
    \mathcal{L}_{init}=\mathcal{L}_c+\mathcal{L}_n+\lambda_{mask}\mathcal{L}_{mask}.
\end{equation}
\textbf{Second stage.} After pretraining the 3D Gaussian model, we freeze geometric Gaussian attributes (positions, covariance, opacity, and normals) and optimize BRDF material parameters jointly with the HDR environment map. We use an L1 photometric loss $\mathcal{L}^{pbr}_1$ with unit weight in this stage. For reproducibility, Tab.~\ref{tab:stagewise_schedule} enumerates trainable/frozen variables and active losses for both stages.

\begin{table}[t]
    \centering
    \caption{Stage-wise optimization schedule (compact form).}
    \label{tab:stagewise_schedule}
    \small
    \setlength{\tabcolsep}{4pt}
    \renewcommand{\arraystretch}{1.05}
    \begin{tabularx}{\linewidth}{lCCC}
        \toprule
        Stage & Trainable variables & Frozen variables & Objective \\
        \midrule
        Stage 1 & $\{\mu,\Sigma,c,\alpha,n\}$ & $\{\rho,r,L_{env}\}$ & $\mathcal{L}_{init}$ \\
        Stage 2 & $\{\rho,r,L_{env}\}$ & $\{\mu,\Sigma,c,\alpha,n\}$ & $\mathcal{L}^{pbr}_1$ \\
        \bottomrule
    \end{tabularx}
    \vspace{2pt}
    \footnotesize Symbols: $\mu$ position, $\Sigma$ covariance, $c$ color, $\alpha$ opacity, $n$ normal, $\rho$ albedo, $r$ roughness, and $L_{env}$ environment-map radiance.
\end{table}

Implementation-specific optimization and runtime settings are summarized in the implementation details paragraph below.

\textbf{Implementation details.} The visibility ray tracer is implemented with NVIDIA OptiX, and our Gaussian-splatting training and relighting pipeline is built on top of the Gaussian-splatting Python framework \textit{GSplat}~\cite{ye:2025:gsplat}. We pretrain for 30,000 iterations in stage 1 and 400 iterations in stage 2. On an NVIDIA RTX A6000 graphics processing unit (GPU), stage 1 takes around 15 minutes and stage 2 around 10 minutes. We observe that longer second-stage training degrades relighting quality because the albedo begins to absorb illumination information. We use a compact $16\times32$ environment map and normal-based hemisphere sampling, which halves the number of rays per pixel compared with full-sphere sampling. With a $16\times32$ environment map, there are 512 candidate directions, after hemisphere culling and empty-space removal, the OptiX visibility pass takes approximately 436\,ms at the evaluation resolution, corresponding to about 127 MRays/s. This measures only transmittance-weighted visibility queries. A full second-stage optimization iteration takes approximately 1.25\,s/it, while final PBR relighting renders at about 0.83\,s/image. Forward relighting spends about 55\% of its time on visibility queries and 44\% on BRDF/environment-map integration.

\begin{table*}[t]
    \centering
    \setlength{\fboxsep}{0pt}
    \begin{tabularx}{\textwidth}{l | >{\centering\arraybackslash}p{1.1cm} | CCC | CCC | CCC}
        \toprule
        \multirow{2}{*}{Method} & 
        \multicolumn{1}{c}{Normal} & \multicolumn{3}{c}{Novel View Synthesis} & \multicolumn{3}{c}{Albedo} & \multicolumn{3}{c}{Relighting} \\
        \cmidrule(lr){2-2}\cmidrule(lr){3-5}\cmidrule(lr){6-8}\cmidrule(lr){9-11}
        & MAE & PSNR & SSIM & LPIPS & PSNR & SSIM & LPIPS & PSNR & SSIM & LPIPS \\
        \cmidrule{1-11}
        NeRFactor & 6.314 & 24.679 & 0.922 & 0.120 & 25.125 & \cellcolor{yellow!20}{0.940} & 0.109 & 23.383 & 0.908 & 0.131  \\
        
        InvRender & \cellcolor{yellow!60}{5.074} & 27.367 & 0.934 & 0.089 & 27.341 & 0.933 & \cellcolor{yellow!20}{0.100} & {23.973} & 0.901 & {0.101} \\
        
        NVDiffrec & {6.078} & {30.696} & {0.962} & \cellcolor{yellow!20}{0.052} & {29.174} & 0.908 & 0.115 & 19.880 & {0.879} & 0.104 \\
        
        GS-IR & \cellcolor{yellow!30}{4.948} &  \cellcolor{orange!30}{35.333} & \cellcolor{orange!30}{0.974} & \cellcolor{red!30}{0.039} & \cellcolor{yellow!60}{30.286} & \cellcolor{yellow!60}{0.941} & \cellcolor{orange!30}{0.084} & \cellcolor{yellow!20}{24.374} & \cellcolor{yellow!20}{0.885} & \cellcolor{yellow!20}{0.096} \\

        TensoIR & \cellcolor{orange!30}{4.100} & \cellcolor{yellow!60}{35.088} & \cellcolor{red!30}{0.976} & \cellcolor{orange!30}{0.040} & \cellcolor{yellow!20}{29.275} & \cellcolor{orange!30}{0.950} & \cellcolor{yellow!60}{0.085} & \cellcolor{yellow!60}{28.581} & \cellcolor{red!30}{0.944} & \cellcolor{yellow!60}{0.081} \\        

        IRGS & \cellcolor{red!30}{3.975} & \cellcolor{red!30}{35.530} & \cellcolor{yellow!20}{0.964} & \cellcolor{yellow!60}{0.049} & \cellcolor{red!30}{33.435} & \cellcolor{red!30}{0.955} & \cellcolor{red!30}{0.076} & \cellcolor{red!30}{30.646} & \cellcolor{yellow!60}{0.935} & \cellcolor{red!30}{0.076} \\        
        \midrule

        Ours & 5.111 & \cellcolor{yellow!20}{34.807} & \cellcolor{yellow!60}{0.967} & 0.053 & \cellcolor{orange!30}{30.305} & {0.934} & {0.101} & \cellcolor{orange!30}{30.538} & \cellcolor{orange!30}{0.941} & \cellcolor{orange!30}{0.077}\\
        \bottomrule
    \end{tabularx}
    \caption{Quantitative comparison on the TensoIR Synthetic dataset \cite{jin:2023:tensoir}. All reported values are aggregated means over the evaluated objects. Most results are taken from the respective papers. For IRGS \cite{gu:2025:irgs}, we ran their evaluation ourselves because the publication did not include the full results.}
    \label{tab:method_comparison}
\end{table*}

\begin{table*}[t]
    \centering
    \setlength{\tabcolsep}{3.2pt}
    \renewcommand{\arraystretch}{1.05}
    \begin{tabularx}{\textwidth}{l | CCC | CCC | CCC}
        \toprule
        \multirow{2}{*}{Method} & 
        \multicolumn{3}{c}{NVS} & \multicolumn{3}{c}{Albedo} & \multicolumn{3}{c}{Relighting} \\
        \cmidrule(lr){2-4}\cmidrule(lr){5-7}\cmidrule(lr){8-10}
        &  PSNR & SSIM & LPIPS & PSNR & SSIM & LPIPS & PSNR & SSIM & LPIPS \\
        \cmidrule{1-10}
        PBR-only & 35.697 & 0.971 & 0.046 & 20.848 & 0.922 & 0.105 & 28.848 & 0.932 & 0.077 \\
        \cmidrule{1-10}
        w/o visibility & 34.789 & 0.966 & 0.053 & 30.313 & 0.933 & 0.101 & 29.091 & 0.933 & 0.077\\
        w/o opacity loss & 34.996 & 0.967 & 0.052 & 29.834 & 0.932 & 0.103 & 29.939 & 0.938 & 0.077 \\        
        \cmidrule{1-10}
        Full & 34.996 & 0.976 & 0.053 & 30.321 & 0.933 & 0.101 & 30.594 & 0.942 & 0.076\\
        \bottomrule
    \end{tabularx}%
    \vspace{0.5mm}
    \caption{Ablation study of the different components of VisibilitySplat on the Lego model from the TensoIR synthetic dataset~\cite{jin:2023:tensoir}.}
    \label{tab:ablation_study}
\end{table*}

\section{Experiments}
We compare our method against previous state-of-the-art methods and also present an ablation study to isolate the contribution of each design component. Baselines are selected based on popularity and code availability.

Our evaluation focuses on masked single-object scenes. This choice narrows the range of real-world inverse-rendering ambiguities (e.g., clutter, contact shadows, interreflections, and segmentation errors), but it is technically motivated by our lighting model: we sample illumination from a distant environment map, and without modeling secondary bounces the incident radiance can vanish when the environment is occluded by surrounding structures. Isolated objects therefore provide a well-defined direct-lighting signal and allow us to isolate the effect of explicit visibility queries.

\subsection{Dataset \& Metrics}
We report experimental results on the synthetic TensoIR dataset \cite{jin:2023:tensoir} for IR of single objects. It contains four objects, each with reference images rendered under ten different lighting conditions, along with albedo and normal maps. We evaluate normal-estimation quality using mean angular error (MAE). For relighting, NVS, and albedo reconstruction, we report peak signal-to-noise ratio (PSNR), structural similarity index (SSIM), and learned perceptual image patch similarity (LPIPS) \cite{zhang:2018:lpips}. Some results of our method for each model can be seen in Fig. \ref{fig:all_models}.

Following prior work, we compensate for the inherent albedo-shading scale ambiguity by scaling each predicted albedo with a single global RGB factor that minimizes MSE to the ground truth \cite{land:1971:lightness, barron:2014:shape, zhang:2021:nerfactor, jin:2023:tensoir, liang:2024:gs-ir, shi:2025:gir, gao:2024:relightable3dgaussians}. We do not apply color correction \cite{buchsbaum:1980:spatial}, all reported and visualized synthetic albedo results therefore use this scale normalization unless stated otherwise. \\
To visualize the HDR environment maps, we follow TensoIR~\cite{jin:2023:tensoir} and apply gamma tonemapping to convert HDR radiance to low-dynamic-range (LDR) values for display. \\
NVS results are rendered using PBR with the estimated material parameters and environment lighting.

\subsection{Comparison}
We compare our method with previous inverse rendering methods, NeRFactor \cite{zhang:2021:nerfactor}, InvRender \cite{zhang:2022:invrender}, NVDiffrec \cite{munkberg:2022:nvdiffrec}, GS-IR \cite{liang:2024:gs-ir}, TensoIR \cite{jin:2023:tensoir} and IRGS \cite{gu:2025:irgs} on the TensoIR synthetic dataset. All comparisons are performed using multi-view images captured under unknown lighting conditions. Tab. \ref{tab:method_comparison} summarizes the quantitative comparison and Fig. \ref{fig:comparison_grid} shows a visual comparison.

Our method is competitive with prior 3DGS-based IR approaches and improves over several earlier baselines across NVS, albedo reconstruction, and relighting. However, IRGS \cite{gu:2025:irgs} achieves the best overall results on this benchmark. These results position VisibilitySplat as a direct explicit-visibility baseline that remains competitive with specialized 3DGS-based IR methods while retaining the standard rasterized 3DGS pipeline for training and visualization. We attribute our performance gains over earlier baselines primarily to the explicit visibility term: on masked, single-object scenes the dominant ambiguity is often cast-shadow visibility and mutual occlusion under direct environment lighting, while multi-bounce transport is comparatively weak. In this regime, sparse but physically grounded transmittance queries already provide a large part of the benefit for disentangling albedo and lighting. Compared to IRGS, our lightweight separation between rasterized splatting for efficient training/preview and sparse ray-based visibility queries offers a favorable practicality-accuracy trade-off, but still leaves room for improvement in terms of absolute reconstruction quality.

Our main source of error is imperfect surface reconstruction rather than instability of the occlusion ray tracer. In particular, inaccurate normals lead to visibly curved shadow boundaries and local relighting artifacts (see Fig. \ref{fig:occlusion_b}). This interpretation is consistent with our higher mean angular error compared to competing methods. Despite these geometric limitations, our relighting quality in PSNR, SSIM, and LPIPS remains competitive with state-of-the-art approaches.

At the measured $800\times800$ evaluation resolution on an NVIDIA RTX A6000 GPU, final PBR relighting takes about 0.83\,s/image, so the current implementation is not real-time. Forward relighting time is dominated by explicit visibility queries (about 55\%) and discrete BRDF/environment-map integration (about 44\%), the remaining time is spent on standard 3D Gaussian splatting and minor overhead. Thus, real-time deployment would require optimizing both the visibility pass and the PBR integration.
\subsection{Ablation Study}

We conduct an ablation study on the components of our method to validate their effectiveness. We report PSNR, SSIM, and LPIPS for NVS, albedo reconstruction, and relighting on the Lego model from the TensoIR Synthetic dataset (Tab. \ref{tab:ablation_study}), including a PBR-only baseline without visibility and without opacity-mask loss, as well as single-component removals.

\textbf{PBR-only baseline.} We disable both the visibility ray tracer and the opacity-mask loss and optimize with PBR supervision alone. Interestingly, this baseline yields the highest NVS PSNR (and lowest LPIPS), though not the highest SSIM. We attribute this behavior to stronger appearance fitting, where albedo absorbs more illumination information and therefore reproduces the training lighting more easily. However, this entanglement hurts decomposition: albedo reconstruction and relighting quality both degrade compared to the full model.

\textbf{Visibility.} To isolate the visibility term, we compare the \textit{without visibility} variant (with opacity-mask loss retained) against the full model. Reintroducing explicit visibility improves relighting metrics and yields cleaner shadow behavior, indicating better separation of material and illumination.

\textbf{Opacity mask loss.} As described in the optimization section, this term mainly acts through improved geometric support for ray-based visibility queries. Comparing \textit{without opacity loss} with the full model shows lower relighting quality. Removing the opacity-mask loss also decreases albedo PSNR, while albedo SSIM and LPIPS remain nearly unchanged. This suggests that coarse albedo structure is preserved but radiometric consistency degrades. Better-defined occluding geometry from opacity regularization allows visibility rays to produce cleaner shadow estimates, which in turn stabilizes relighting.

\section{Conclusion}

We present VisibilitySplat, an IR approach that augments 3DGS with a sparse visibility oracle for transmittance-weighted occlusion. Implemented using BVH-accelerated point-based ray tracing, the oracle complements the standard rasterized 3DGS training and visualization loop while injecting explicit visibility cues for material-illumination disentanglement. On a synthetic benchmark, we improve over several baselines in PSNR for NVS, albedo reconstruction, and relighting, while remaining competitive in SSIM and LPIPS. Ablations suggest that explicit visibility queries primarily improve relighting and decomposition quality, even when a PBR-only variant can score higher on NVS appearance metrics.

\textbf{Limitations.} Our evaluation targets single, masked, static objects, behavior in cluttered or dynamic scenes, or under imperfect segmentation, is not assessed. The illumination model accounts for direct lighting only. Indirect transport (interreflection, subsurface, participating media) and very high-frequency environment lighting are not modeled. Finally, experiments are restricted to synthetic data, robustness to real captures, complex materials (e.g., glass, hair, translucency), and sensor noise remains untested.

\textbf{Future work.} Future extensions include multi-object and full-scene evaluation, incorporation of indirect lighting, tighter coupling between splatting and visibility (e.g., differentiable or cached variants), improved geometric robustness, and investigations toward real-time deployment and comprehensive testing on real datasets.

\textbf{Code availability.} We will release the full implementation and supporting resources upon publication of this paper.

\bibliographystyle{plain}
\bibliography{main_bibliography}

\begin{thebibliography}{10}

\bibitem{barron:2014:shape}
Jonathan~T Barron and Jitendra Malik.
\newblock Shape, illumination, and reflectance from shading.
\newblock {\em IEEE transactions on pattern analysis and machine intelligence},
  37(8):1670--1687, 2014.

\bibitem{barron:2021:mip}
Jonathan~T Barron, Ben Mildenhall, Matthew Tancik, Peter Hedman, Ricardo
  Martin-Brualla, and Pratul~P Srinivasan.
\newblock {Mip-NeRF}: A multiscale representation for anti-aliasing neural
  radiance fields.
\newblock In {\em Proceedings of the IEEE/CVF international conference on
  computer vision}, pages 5855--5864, 2021.

\bibitem{barron:2022:mip360}
Jonathan~T Barron, Ben Mildenhall, Dor Verbin, Pratul~P Srinivasan, and Peter
  Hedman.
\newblock {Mip-NeRF 360}: Unbounded anti-aliased neural radiance fields.
\newblock In {\em Proceedings of the IEEE/CVF conference on computer vision and
  pattern recognition}, pages 5470--5479, 2022.

\bibitem{barron:2023:zip}
Jonathan~T Barron, Ben Mildenhall, Dor Verbin, Pratul~P Srinivasan, and Peter
  Hedman.
\newblock {Zip-NeRF}: Anti-aliased grid-based neural radiance fields.
\newblock In {\em Proceedings of the IEEE/CVF International Conference on
  Computer Vision}, pages 19697--19705, 2023.

\bibitem{buchsbaum:1980:spatial}
Gershon Buchsbaum.
\newblock A spatial processor model for object colour perception.
\newblock {\em Journal of the Franklin institute}, 310(1):1--26, 1980.

\bibitem{burley:2012:physically}
Brent Burley.
\newblock Physically-based shading at {D}isney.
\newblock In {\em ACM SIGGRAPH 2012 Courses: Practical Physically-Based Shading
  in Film and Game Production}, pages 1--7, 2012.

\bibitem{chen:2022:tensorf}
Anpei Chen, Zexiang Xu, Andreas Geiger, Jingyi Yu, and Hao Su.
\newblock {TensoRF}: Tensorial radiance fields.
\newblock In {\em European conference on computer vision}, pages 333--350.
  Springer, 2022.

\bibitem{chen:2024:gi-gs}
Hongze Chen, Zehong Lin, and Jun Zhang.
\newblock {GI-GS}: Global illumination decomposition on gaussian splatting for
  inverse rendering.
\newblock {\em arXiv preprint arXiv:2410.02619}, 2024.

\bibitem{debevec:1998:rendering}
Paul Debevec.
\newblock Rendering synthetic objects into real scenes: bridging traditional
  and image-based graphics with global illumination and high dynamic range
  photography.
\newblock In {\em Proceedings of the 25th Annual Conference on Computer
  Graphics and Interactive Techniques}, SIGGRAPH '98, pages 189--198, New York,
  NY, USA, 1998. Association for Computing Machinery.

\bibitem{dihlmann:2024:subsurface}
Jan-Niklas Dihlmann, Arjun Majumdar, Andreas Engelhardt, Raphael Braun, and
  Hendrik Lensch.
\newblock Subsurface scattering for gaussian splatting.
\newblock {\em Advances in Neural Information Processing Systems},
  37:121765--121789, 2024.

\bibitem{fan:2025:rng}
Jiahui Fan, Fujun Luan, Jian Yang, Milos Hasan, and Beibei Wang.
\newblock {RNG}: Relightable neural gaussians.
\newblock In {\em Proceedings of the Computer Vision and Pattern Recognition
  Conference}, pages 26525--26534, 2025.

\bibitem{fridovich:2023:kplanes}
Sara Fridovich-Keil, Giacomo Meanti, Frederik~Rahb{\ae}k Warburg, Benjamin
  Recht, and Angjoo Kanazawa.
\newblock K-planes: Explicit radiance fields in space, time, and appearance.
\newblock In {\em Proceedings of the IEEE/CVF Conference on Computer Vision and
  Pattern Recognition}, pages 12479--12488, 2023.

\bibitem{fridovich:2022:plenoxels}
Sara Fridovich-Keil, Alex Yu, Matthew Tancik, Qinhong Chen, Benjamin Recht, and
  Angjoo Kanazawa.
\newblock Plenoxels: Radiance fields without neural networks.
\newblock In {\em Proceedings of the IEEE/CVF conference on computer vision and
  pattern recognition}, pages 5501--5510, 2022.

\bibitem{gao:2024:relightable3dgaussians}
Jian Gao, Chun Gu, Youtian Lin, Zhihao Li, Hao Zhu, Xun Cao, Li~Zhang, and Yao
  Yao.
\newblock Relightable {3D} gaussians: Realistic point cloud relighting with
  brdf decomposition and ray tracing.
\newblock In {\em European Conference on Computer Vision}, pages 73--89.
  Springer, 2024.

\bibitem{gu:2025:irgs}
Chun Gu, Xiaofei Wei, Zixuan Zeng, Yuxuan Yao, and Li~Zhang.
\newblock {IRGS}: Inter-reflective gaussian splatting with 2d gaussian ray
  tracing.
\newblock In {\em Proceedings of the Computer Vision and Pattern Recognition
  Conference}, pages 10943--10952, 2025.

\bibitem{huang:2024:2dgs}
Binbin Huang, Zehao Yu, Anpei Chen, Andreas Geiger, and Shenghua Gao.
\newblock {2D} gaussian splatting for geometrically accurate radiance fields.
\newblock In {\em ACM SIGGRAPH 2024 conference papers}, pages 1--11, 2024.

\bibitem{jiang:2024:gaussianshader}
Yingwenqi Jiang, Jiadong Tu, Yuan Liu, Xifeng Gao, Xiaoxiao Long, Wenping Wang,
  and Yuexin Ma.
\newblock {GaussianShader}: {3D} gaussian splatting with shading functions for
  reflective surfaces.
\newblock In {\em Proceedings of the IEEE/CVF Conference on Computer Vision and
  Pattern Recognition}, pages 5322--5332, 2024.

\bibitem{jin:2023:tensoir}
Haian Jin, Isabella Liu, Peijia Xu, Xiaoshuai Zhang, Songfang Han, Sai Bi,
  Xiaowei Zhou, Zexiang Xu, and Hao Su.
\newblock {TensoIR}: Tensorial inverse rendering.
\newblock In {\em Proceedings of the IEEE/CVF Conference on Computer Vision and
  Pattern Recognition}, pages 165--174, 2023.

\bibitem{kerbl:2023:3dgs}
Bernhard Kerbl, Georgios Kopanas, Thomas Leimk{\"u}hler, and George Drettakis.
\newblock 3d gaussian splatting for real-time radiance field rendering.
\newblock {\em ACM Transactions on Graphics}, 42(4), July 2023.

\bibitem{land:1971:lightness}
Edwin~H Land and John~J McCann.
\newblock Lightness and retinex theory.
\newblock {\em Journal of the Optical society of America}, 61(1):1--11, 1971.

\bibitem{liang:2023:envidr}
Ruofan Liang, Huiting Chen, Chunlin Li, Fan Chen, Selvakumar Panneer, and
  Nandita Vijaykumar.
\newblock {ENVIDR}: Implicit differentiable renderer with neural environment
  lighting.
\newblock In {\em Proceedings of the IEEE/CVF International Conference on
  Computer Vision}, pages 79--89, 2023.

\bibitem{liang:2024:gs-ir}
Zhihao Liang, Qi~Zhang, Ying Feng, Ying Shan, and Kui Jia.
\newblock {GS-IR}: {3D} gaussian splatting for inverse rendering.
\newblock In {\em Proceedings of the IEEE/CVF Conference on Computer Vision and
  Pattern Recognition}, pages 21644--21653, 2024.

\bibitem{ling:2024:nerf-as-emitter}
Jingwang Ling, Ruihan Yu, Feng Xu, Chun Du, and Shuang Zhao.
\newblock {NeRF} as a non-distant environment emitter in physics-based inverse
  rendering.
\newblock In {\em ACM SIGGRAPH 2024 Conference Papers}, pages 1--12, 2024.

\bibitem{liu:2024:bigs}
Zhenyuan Liu, Yu~Guo, Xinyuan Li, Bernd Bickel, and Ran Zhang.
\newblock {BiGS}: Bidirectional gaussian primitives for relightable 3d gaussian
  splatting.
\newblock {\em arXiv preprint arXiv:2408.13370}, 2024.

\bibitem{mildenhall:2021:nerf}
Ben Mildenhall, Pratul~P Srinivasan, Matthew Tancik, Jonathan~T Barron, Ravi
  Ramamoorthi, and Ren Ng.
\newblock {NeRF}: Representing scenes as neural radiance fields for view
  synthesis.
\newblock {\em Communications of the ACM}, 65(1):99--106, 2021.

\bibitem{loccoz:2024:3dgrt}
Nicolas Moenne-Loccoz, Ashkan Mirzaei, Or~Perel, Riccardo de~Lutio,
  Janick~Martinez Esturo, Gavriel State, Sanja Fidler, Nicholas Sharp, and Zan
  Gojcic.
\newblock 3d gaussian ray tracing: Fast tracing of particle scenes.
\newblock {\em ACM Transactions on Graphics and SIGGRAPH Asia}, 2024.

\bibitem{muller:2022:instantngp}
Thomas M{\"u}ller, Alex Evans, Christoph Schied, and Alexander Keller.
\newblock Instant neural graphics primitives with a multiresolution hash
  encoding.
\newblock {\em ACM transactions on graphics (TOG)}, 41(4):1--15, 2022.

\bibitem{munkberg:2022:nvdiffrec}
Jacob Munkberg, Jon Hasselgren, Tianchang Shen, Jun Gao, Wenzheng Chen, Alex
  Evans, Thomas M\"uller, and Sanja Fidler.
\newblock {Extracting Triangular {3D} Models, Materials, and Lighting From
  Images}.
\newblock In {\em Proceedings of the IEEE/CVF Conference on Computer Vision and
  Pattern Recognition (CVPR)}, pages 8280--8290, June 2022.

\bibitem{ramamoorthi:2001:AnER}
Ravi Ramamoorthi and Pat Hanrahan.
\newblock An efficient representation for irradiance environment maps.
\newblock {\em Proceedings of the 28th annual conference on Computer graphics
  and interactive techniques}, 2001.

\bibitem{shi:2025:gir}
Yahao Shi, Yanmin Wu, Chenming Wu, Xing Liu, Chen Zhao, Haocheng Feng, Jian
  Zhang, Bin Zhou, Errui Ding, and Jingdong Wang.
\newblock {GIR}: {3D} gaussian inverse rendering for relightable scene
  factorization.
\newblock {\em IEEE Transactions on Pattern Analysis and Machine Intelligence},
  2025.

\bibitem{verbin:2022:ref}
Dor Verbin, Peter Hedman, Ben Mildenhall, Todd Zickler, Jonathan~T Barron, and
  Pratul~P Srinivasan.
\newblock {Ref-NeRF}: Structured view-dependent appearance for neural radiance
  fields.
\newblock In {\em 2022 IEEE/CVF Conference on Computer Vision and Pattern
  Recognition (CVPR)}, pages 5481--5490. IEEE, 2022.

\bibitem{wang:2021:neus}
Peng Wang, Lingjie Liu, Yuan Liu, Christian Theobalt, Taku Komura, and Wenping
  Wang.
\newblock {NeuS}: Learning neural implicit surfaces by volume rendering for
  multi-view reconstruction.
\newblock {\em arXiv preprint arXiv:2106.10689}, 2021.

\bibitem{wu:2024:deferredgs}
Tong Wu, Jia-Mu Sun, Yu-Kun Lai, Yuewen Ma, Leif Kobbelt, and Lin Gao.
\newblock {DeferredGS}: Decoupled and editable gaussian splatting with deferred
  shading.
\newblock {\em arXiv preprint arXiv:2404.09412}, 2024.

\bibitem{yao:2024:reflective}
Yuxuan Yao, Zixuan Zeng, Chun Gu, Xiatian Zhu, and Li~Zhang.
\newblock Reflective gaussian splatting.
\newblock {\em arXiv preprint arXiv:2412.19282}, 2024.

\bibitem{yariv:2021:volsdf}
Lior Yariv, Jiatao Gu, Yoni Kasten, and Yaron Lipman.
\newblock Volume rendering of neural implicit surfaces.
\newblock {\em Advances in neural information processing systems},
  34:4805--4815, 2021.

\bibitem{ye:2025:gsplat}
Vickie Ye, Ruilong Li, Justin Kerr, et~al.
\newblock {GSplat}: An open-source library for gaussian splatting.
\newblock {\em Journal of Machine Learning Research}, 26(34):1--17, 2025.

\bibitem{yu:2024:gaussian}
Zehao Yu, Torsten Sattler, and Andreas Geiger.
\newblock Gaussian opacity fields: Efficient adaptive surface reconstruction in
  unbounded scenes.
\newblock {\em ACM Transactions on Graphics (ToG)}, 43(6):1--13, 2024.

\bibitem{zhang:2023:neilf++}
Jingyang Zhang, Yao Yao, Shiwei Li, Jingbo Liu, Tian Fang, David McKinnon,
  Yanghai Tsin, and Long Quan.
\newblock {NeILF++}: Inter-reflectable light fields for geometry and material
  estimation.
\newblock In {\em Proceedings of the IEEE/CVF International Conference on
  Computer Vision}, pages 3601--3610, 2023.

\bibitem{zhang:2021:physg}
Kai Zhang, Fujun Luan, Qianqian Wang, Kavita Bala, and Noah Snavely.
\newblock {PhySG}: Inverse rendering with spherical gaussians for physics-based
  material editing and relighting.
\newblock In {\em Proceedings of the IEEE/CVF conference on computer vision and
  pattern recognition}, pages 5453--5462, 2021.

\bibitem{zhang:2018:lpips}
Richard Zhang, Phillip Isola, Alexei~A Efros, Eli Shechtman, and Oliver Wang.
\newblock The unreasonable effectiveness of deep features as a perceptual
  metric.
\newblock In {\em CVPR}, 2018.

\bibitem{zhang:2021:nerfactor}
Xiuming Zhang, Pratul~P Srinivasan, Boyang Deng, Paul Debevec, William~T
  Freeman, and Jonathan~T Barron.
\newblock {NeRFactor}: Neural factorization of shape and reflectance under an
  unknown illumination.
\newblock {\em ACM Transactions on Graphics (ToG)}, 40(6):1--18, 2021.

\bibitem{zhang:2022:invrender}
Yuanqing Zhang, Jiaming Sun, Xingyi He, Huan Fu, Rongfei Jia, and Xiaowei Zhou.
\newblock Modeling indirect illumination for inverse rendering.
\newblock In {\em Proceedings of the IEEE/CVF Conference on Computer Vision and
  Pattern Recognition}, pages 18643--18652, 2022.

\bibitem{zwicker:2002:ewa}
Matthias Zwicker, Hanspeter Pfister, Jeroen Van~Baar, and Markus Gross.
\newblock {EWA} splatting.
\newblock {\em IEEE Transactions on Visualization and Computer Graphics},
  8(3):223--238, 2002.

\end{thebibliography}

\end{document}